%% file: main.tex
\pdfoutput=1
\PassOptionsToPackage{table}{xcolor} %
\PassOptionsToPackage{hyperfootnotes=false}{hyperref} %
\PassOptionsToPackage{patch={item,toc,footnote}}{microtype} %
\documentclass[11pt, a4paper, copyright, google]{google}

\usepackage[authoryear, sort&compress, round]{natbib}
\input{math_commands.tex} %
\usepackage{footmisc} %
\usepackage{xspace}
\usepackage{subcaption}
\usepackage{wrapfig}
\usepackage{silence}
\usepackage[htt]{hyphenat} %
\usepackage{etoolbox}

\definecolor{maincolor}{HTML}{005CAB}
\colorlet{mainrow}{maincolor!12} %
\hypersetup{colorlinks=true,linkcolor=maincolor,citecolor=maincolor,urlcolor=maincolor}
\definecolor{rankone}{HTML}{B5DDB8}
\definecolor{ranktwo}{HTML}{D6ECD8}
\definecolor{rankrest}{HTML}{EEF6EE}

\renewcommand{\arraystretch}{1.05}
\DeclareCaptionLabelSeparator{pipe}{~$\vert$~} %

\newcommand{\grounded}[2]{#2} %
\newif\ifsysstyled
\sysstyledfalse %
\newcommand{\sysplain}{DuoDiscover}
\newcommand{\sysstyledform}{\textbf{DuoDiscover}\null} %
\newcommand{\sys}{\texorpdfstring{\ifsysstyled\sysstyledform\else\sysplain\fi}{\sysplain}\xspace}
\newcommand{\syshl}{\texorpdfstring{\textcolor{maincolor}{\textbf{DuoDiscover}}\null}{\sysplain}\xspace}
\newif\ifsyslogointables
\syslogointablesfalse %
\newcommand{\syslogomark}{\ifsyslogointables\raisebox{-0.18ex}{\includegraphics[height=0.85em]{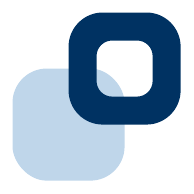}}\hspace{0.22em}\fi}
\newcommand{\systab}{\syslogomark\sys}

\newcommand{\solexec}{SOL-Exec\xspace}
\newcommand{\jax}{JAXBench\xspace}
\newcommand{\tpuinf}{TPU-Inference\xspace}
\newcommand{\fa}{FA4\xspace}
\newcommand{\formulacode}{FormulaCode\xspace}
\newcommand{\sweeff}{SWE-fficiency\xspace}
\newcommand{\claudecode}{Claude Code\xspace}
\newcommand{\openevolve}{OpenEvolve\xspace}
\newcommand{\openevolveagentic}{OpenEvolve-Agentic\xspace}
\newcommand{\avo}{AVO\xspace}
\newcommand{\kda}{KDA\xspace}

\newcommand{\perfagent}{PerfAgent\xspace}
\newcommand{\xspeedup}[2]{#1 to #2$\times$}
\newcommand{\diffspeedup}[1]{#1$\times$}

\definecolor{shubhamcol}{rgb}{0.60,0.20,0.65}
\newcommand{\shubham}[1]{{\color{shubhamcol}\textsf{\scriptsize /*~Shubham:~#1~*/}}}
\newcommand{\cl}[1]{{\color{olive}\textsf{\scriptsize /*~CL:~#1~*/}}}
\newcommand{\dilara}[1]{{\color{violet}\textsf{\scriptsize /*~Dilara:~#1~*/}}}
\newcommand{\mert}[1]{{\color{magenta}\textsf{\scriptsize /*~Mert:~#1~*/}}}
\newcommand{\violet}[1]{{\color{purple}\itshape\scriptsize /*~Violet:~#1~*/}} %
\newcommand{\jiefengc}[1]{{\color{blue}\textsf{\scriptsize /*~Jiefeng:~#1~*/}}}
\definecolor{bhavanacol}{rgb}{0.90,0.30,0.90}
\newcommand{\bhavana}[1]{{\color{bhavanacol}\textsf{\scriptsize /*~Bhavana:~#1~*/}}}
\newcommand{\mike}[1]{{\color{orange}\textsf{\scriptsize /*~Mike:~#1~*/}}}
\definecolor{akshatcol}{rgb}{0.10,0.50,0.80}
\newcommand{\akshat}[1]{{\color{akshatcol}\textsf{\small /* ~ Akshat: ~ #1 ~ */}}}
\definecolor{ioncol}{rgb}{0.55,0.27,0.07}
\newcommand{\ion}[1]{{\color{ioncol}\textsf{\scriptsize /*~Ion:~#1~*/}}}
\definecolor{patcol}{rgb}{0.0,0.0,0.5}
\newcommand{\pat}[1]{{\color{patcol}\textsf{\scriptsize /*~Pat:~#1~*/}}}
\definecolor{abecol}{rgb}{0.0,0.55,0.45}
\newcommand{\abe}[1]{{\color{abecol}\textsf{\scriptsize /*~Abe:~#1~*/}}}
\newif\ifshowcomments
\showcommentstrue
\ifshowcomments\else
  \renewcommand{\shubham}[1]{}\renewcommand{\cl}[1]{}\renewcommand{\dilara}[1]{}\renewcommand{\mert}[1]{}
  \renewcommand{\violet}[1]{}\renewcommand{\jiefengc}[1]{}\renewcommand{\bhavana}[1]{}\renewcommand{\mike}[1]{}
  \renewcommand{\akshat}[1]{}\renewcommand{\ion}[1]{}\renewcommand{\pat}[1]{}\renewcommand{\abe}[1]{}
\fi
\definecolor{agentcol}{HTML}{B8860B} %

\definecolor{agentbcol}{rgb}{0.85,0.10,0.55}

\definecolor{agentrcol}{rgb}{0.85,0.10,0.10}

\definecolor{agentgcol}{rgb}{0.00,0.50,0.15}

\AddToHook{para/begin}{\looseness=-1}
\AtBeginEnvironment{thebibliography}{\hbadness=9999\emergencystretch=3em} %
\makeatletter\input{t1zlmtt.fd}\makeatother %
\DeclareFontShape{T1}{zlmtt}{b}{sc}{<->ssub*zlmtt/b/n}{}
\DeclareFontShape{T1}{zlmtt}{b}{scsl}{<->ssub*zlmtt/b/sc}{}
\DeclareFontShape{T1}{zlmtt}{bx}{sc}{<->ssub*zlmtt/bx/n}{}
\DeclareFontShape{T1}{zlmtt}{bx}{scsl}{<->ssub*zlmtt/bx/sc}{}
\DeclareFontShape{T1}{zlmtt}{bx}{ui}{<->ssub*zlmtt/bx/n}{}
\DeclareFontShape{T1}{zlmtt}{c}{it}{<->ssub*zlmtt/c/n}{}
\DeclareFontShape{T1}{zlmtt}{c}{n}{<->ssub*zlmtt/m/n}{}
\DeclareFontShape{T1}{zlmtt}{c}{sc}{<->ssub*zlmtt/c/n}{}
\DeclareFontShape{T1}{zlmtt}{c}{scit}{<->ssub*zlmtt/c/sc}{}
\DeclareFontShape{T1}{zlmtt}{c}{sl}{<->ssub*zlmtt/c/it}{}
\DeclareFontShape{T1}{zlmtt}{lc}{sc}{<->ssub*zlmtt/lc/n}{}
\DeclareFontShape{T1}{zlmtt}{lc}{scsl}{<->ssub*zlmtt/lc/sc}{}
\DeclareFontShape{T1}{zlmtt}{m}{In}{<->ssub*zlmtt/m/n}{}
\DeclareFontShape{T1}{zlmtt}{m}{ui}{<->ssub*zlmtt/m/n}{}
\DeclareFontShape{T1}{zlmtt}{sb}{it}{<->ssub*zlmtt/sb/n}{}
\DeclareFontShape{T1}{zlmtt}{sb}{n}{<->ssub*zlmtt/m/n}{}
\DeclareFontShape{T1}{zlmtt}{sb}{sc}{<->ssub*zlmtt/sb/n}{}
\DeclareFontShape{T1}{zlmtt}{sb}{scit}{<->ssub*zlmtt/sb/sc}{}
\DeclareFontShape{T1}{zlmtt}{sb}{sl}{<->ssub*zlmtt/sb/it}{}
\DeclareFontShape{T1}{zlmtt}{sbc}{sc}{<->ssub*zlmtt/c/sc}{}
\DeclareFontShape{T1}{zlmtt}{sbx}{sc}{<->ssub*zlmtt/sb/sc}{}
\DeclareFontShape{T1}{zlmtt}{sbx}{scsl}{<->ssub*zlmtt/sb/scsl}{}
\DeclareFontShape{T1}{zlmtt}{sb}{scsl}{<->ssub*zlmtt/sb/scit}{}
\DeclareFontShape{T1}{zlmtt}{sbc}{scit}{<->ssub*zlmtt/c/scit}{}
\DeclareFontShape{T1}{zlmtt}{c}{scsl}{<->ssub*zlmtt/c/scit}{}
\DeclareFontShape{T1}{zlmtt}{sbx}{n}{<->ssub*zlmtt/m/n}{}
\DeclareFontShape{T1}{zlmtt}{sbx}{it}{<->ssub*zlmtt/sbx/n}{}
\DeclareFontShape{T1}{zlmtt}{sbx}{sl}{<->ssub*zlmtt/sbx/it}{}
\DeclareFontShape{T1}{zlmtt}{sbx}{scit}{<->ssub*zlmtt/sbx/scsl}{}
\DeclareFontShape{T1}{zlmtt}{sbc}{scsl}{<->ssub*zlmtt/c/scsl}{}
\DeclareFontShape{T1}{zlmtt}{sbc}{n}{<->ssub*zlmtt/m/n}{}
\DeclareFontShape{T1}{zlmtt}{sbc}{it}{<->ssub*zlmtt/sbc/n}{}
\DeclareFontShape{T1}{zlmtt}{sbc}{sl}{<->ssub*zlmtt/sbc/it}{}

\makeatletter
\newif\if@pagehasfn
\renewcommand{\footnoterule}{\global\@pagehasfntrue\kern-1pt{\color{gray}\hrule width\columnwidth height 1pt}\kern 5pt}
\AddToHook{shipout/after}{\global\@pagehasfnfalse}
\renewcommand{\footrule}{\if@pagehasfn\else{\color{gray}\DefaultFootRule}\fi}
\makeatother

\renewcommand{\titlefont}{\color{black}\normalfont\bfseries\fontsize{19}{21}\selectfont}
\uselogo{}
\newif\iftitlelogodone
\newcommand{\titlelogo}{\iftitlelogodone\else\raisebox{-0.25ex}{\includegraphics[height=1.05em]{assets/logo_duodiscover}}\hspace{0.35em}\global\titlelogodonetrue\fi}

\renewcommand\Affilfont{\normalfont\fontsize{9}{11}\selectfont} %
\makeatletter
\patchcmd{\maketitle}{{\raggedright \@author\par}}{{\rightskip-24pt\relax\@author\par}}{}{\PackageWarning{main}{author-justify patch failed}}
\patchcmd{\maketitle}{\vskip20pt}{\vskip12pt}{}{\PackageWarning{main}{authors-abstract gap patch failed}} %
\renewcommand\AB@affilsepx{\hspace{0.75em}\protect\Affilfont} %
\makeatother
\renewcommand{\today}{2026-10-12} %

\title{\texorpdfstring{\titlelogo}{}\syshl: Scaling Autonomous~Discovery Beyond the Evaluation Bottleneck}
\author[*1,2]{Shubham~Agarwal}
\author[*1,2]{Akshat~Gupta}
\author[*1]{Mike~Heddes}
\author[*1,3]{Dilara~Soylu}
\author[1]{,~Rui~Meng} %
\author[1]{Patrick~Tser~Jern~Kon}
\author[1]{Jiefeng~Chen}
\author[1,2]{Mert~Cemri}
\author[2]{Shu~Liu}
\author[1]{Abe~Gonzalez}
\author[1]{Akanksha~Jain}
\author[1]{Bhavana~Dalvi~Mishra}
\author[1]{Linfeng~Song}
\author[1]{Vilobh~Meshram}
\author[1]{Xiangru~Tang}
\author[1]{Jinsung~Yoon}
\author[2]{Aditya~Parameswaran}
\author[2]{Ion~Stoica}
\author[1]{Vahab~Mirrokni}
\author[1]{Burak~Gokturk}
\author[1]{Parthasarathy~Ranganathan}
\author[1]{Tomas~Pfister}
\author[$\dagger$1]{Nanyun~Peng}
\author[$\dagger$1]{Chun-Liang~Li}
\makeatletter\@namedef{@sep5}{}\makeatother

\affil[1]{Google}
\affil[2]{UC Berkeley}
\affil[3]{Stanford University}
\makeatletter\gdef\AB@affilsep{\protect\\[1pt]}\makeatother
\affil[*]{Equal contribution, ordered alphabetically}
\makeatletter\gdef\AB@affilsep{. \protect\Affilfont}\makeatother
\affil[$\dagger$]{Co-last authors.}

\begin{abstract}
Evaluation has been central to advances in AI, and it is equally critical to autonomous discovery. Yet scaling autonomous discovery to real-world engineering challenges faces a fundamental bottleneck: the cost of evaluating progress.
In real settings, a single true evaluation, such as a full benchmark or serving trace, can take hours, so a search can afford only a handful of true evaluations.
We present DuoDiscover, a search system that jointly discovers what to optimize and how to evaluate progress efficiently, through three mechanisms.
First, each worker in a fleet of coding agents co-discovers an optimization target and a fast, tailored proxy evaluator, which we call a \emph{yardstick}.
The agent then hill-climbs the yardstick rather than the true evaluation.
Second, since search can exploit gaps between a yardstick and the true objective, fidelity tests continually recheck each yardstick's ranking and invalidate any that drift.
Third, the fleet is bottom-up, with no central decomposition of the problem. A lightweight orchestrator spawns and terminates agents, briefs running ones, and seeds new agents with others' confirmed work.
DuoDiscover enables autonomous optimization from software to chip architecture. 
We demonstrate DuoDiscover on two repository-level benchmarks and four GPU and TPU kernel suites covering hundreds of kernels.
On SWE-fficiency, it doubles the average speedup of human experts. On SOL-Exec, it ranks first on 121 of 235 kernels and in the top three on 212.
On the heavily optimized vLLM and SQLite engines, DuoDiscover achieves 1.92$\times$ and 1.38$\times$ gains over the base code. On a CPU microarchitecture optimization task with evaluations of up to 28 hours, DuoDiscover surpasses both AlphaEvolve-based ArchAgent v2 and the best human design, with 1.65$\times$ ArchAgent v2's gain in 16.7 days versus more than 40 days, advancing the frontier of autonomous chip design.
\end{abstract}

\begin{document}
{\hbadness=10000\hfuzz=2.5pt\maketitle\par}
\vspace{10pt}\noindent\begin{minipage}{\linewidth}
\centering
\setlength{\abovecaptionskip}{4pt}
\captionsetup{hypcap=false} %
\includegraphics[width=0.92\linewidth]{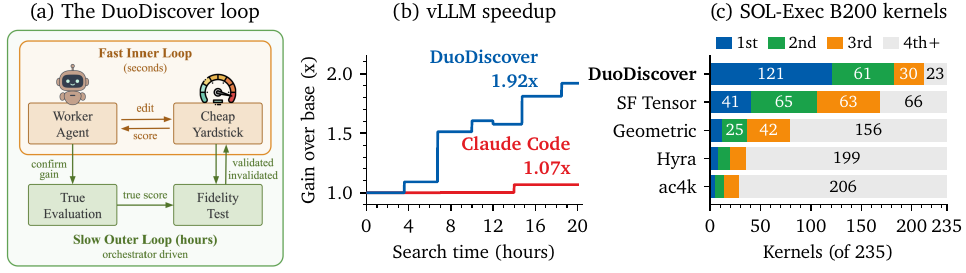}
\captionof{figure}{\sys at a glance.
(a) A worker agent hill-climbs a cheap \emph{yardstick} in an inner loop (seconds per edit), while an outer loop of true evaluations (hours each) confirms gains and runs the fidelity test.
(b) On vLLM, \sys reaches 1.92$\times$ on a four-objective geometric mean, against Claude Code's 1.07$\times$.
(c) On NVIDIA's \solexec leaderboard, \sys takes first place. }
\label{fig:teaser}
\end{minipage}

\section{Introduction}

Search-based discovery systems that couple large language models (LLMs) with iterative search and evaluator feedback have achieved strong results across mathematics \citep{romera2024funsearch}, algorithms \citep{novikov2025alphaevolve}, and systems optimization \citep{cheng2025barbarians}. 
Evaluation has long driven progress in AI, yet most prior work on autonomous discovery focuses on refining the search harness itself \citep{lange2026shinkaevolve, cemri2026adaevolve}, largely treating the evaluator as given and fixed. 
On real engineering systems, however, that assumption breaks down: evaluating progress becomes the bottleneck.
For instance, optimizing vLLM for multi-objective serving requires 85 minutes per true evaluation, limiting even frontier coding agents to a dozen evaluations within a 20-hour search budget.
Overcoming this bottleneck is key to scaling autonomous discovery across science and real-world engineering.

A natural solution is to guide search with cheap proxy evaluators instead of expensive true evaluations. This works because large speedups often come from focused edits to a few small regions of code, which we call \emph{targets}, whose progress can be measured much more cheaply than the entire true evaluation. However, proxy evaluators are usually target-specific, making it impractical to design one by hand for every target.
Moreover, effective proxy-guided search requires jointly identifying promising targets and cheap, faithful proxies: optimizing the wrong target yields little benefit, while an inaccurate proxy can misguide search. Even faithful proxies can become unreliable as search exploits discrepancies between the proxy and the true objective, improving proxy scores without real gains.

We propose \syshl, a search system that jointly discovers what to optimize and how to evaluate progress when evaluation becomes the bottleneck.
Each worker agent co-discovers a promising optimization target and a cheap proxy evaluator, called a \emph{yardstick}, then hill-climbs on the yardstick to find better solutions. 
Beyond co-discovery, \sys addresses two further challenges. 
The first is maintaining \emph{evaluation fidelity}.
A yardstick that initially ranks edits like the true objective can drift as the agent adapts to it. \sys continually runs fidelity tests to invalidate drifting yardsticks and requires a true evaluation to confirm every gain before it is accepted. 
The second is \emph{target selection}, choosing which regions to optimize among many candidate targets with unpredictable payoffs. Rather than decomposing the problem centrally, \sys deploys a bottom-up fleet of independent worker agents for co-discovery. Because worker agents are cheap but true evaluations are expensive, a lightweight orchestrator manages only the true evaluations: it steers them toward agents making progress, terminates stalled agents, and combines compatible gains.

\sys delivers autonomous optimization at every level of the computing stack, from software to chip architecture. 
We demonstrate it on two repository benchmarks, four GPU/TPU kernel suites, and real-world systems. 
On repositories, \sys doubles the human-expert speedup on SWE-fficiency and beats all baselines on \formulacode, by 21\% over the strongest. For GPU kernels, \sys achieves a 14\% higher speed-of-light (SOL) score than the best baseline, ranking first on \solexec for 121 of 235 kernels and in the top three for 212 at the time of submission. On \tpuinf kernels, \sys attains a \grounded{tpuinf.duodiscover}{2.68}$\times$ speedup over optimized Pallas kernels, ahead of the best baseline at \grounded{tpuinf.claudecode}{2.38}$\times$.
On JAXBench, \sys achieves a \grounded{jax.duodiscover}{4.62}$\times$ speedup over XLA. On highly optimized systems (vLLM and SQLite), \sys finds up to a 1.92$\times$ speedup. 
Finally, on an expensive CPU microarchitecture design task, whose evaluation takes up to 28 hours, it surpasses both the AlphaEvolve-based ArchAgent v2 and the best human design, with 1.65$\times$ ArchAgent v2's gain using about 8{,}000 candidates over 16.7 days, against over 12{,}000 candidates and more than 40 days for ArchAgent v2, advancing the frontier of autonomous chip design.

\section{\sys: Co-discovering targets and yardsticks}\label{sec:method}

\begin{figure}[t]
\centering
\includegraphics[width=\linewidth]{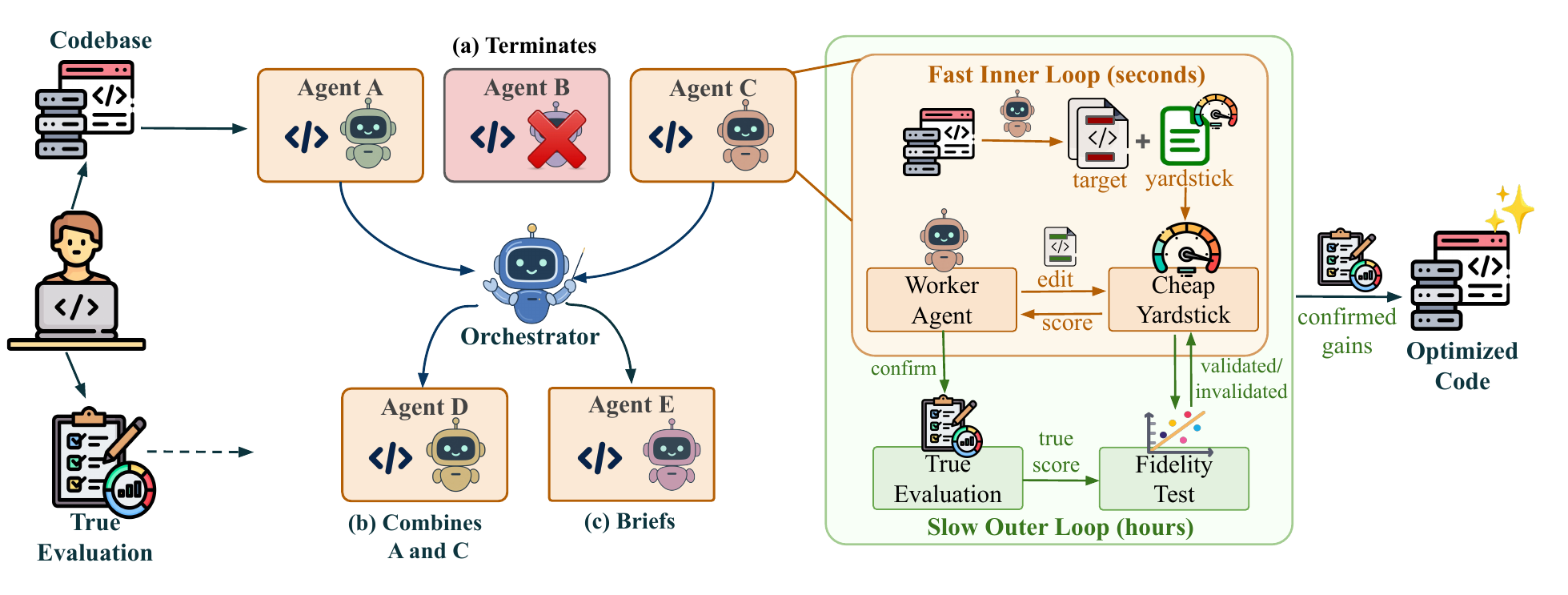}
\vspace{-2em}
\caption{Overview of \sys. Left: a fleet of worker agents, each on its own copy of the codebase, coordinated by an orchestrator that terminates agents (a), spawns new ones, here one combining A and C (b), and briefs running ones (c). Right: inside one agent, a fast loop co-discovers a target and yardstick and climbs it on cheap scores, while a slow loop confirms gains with true evaluations and runs the fidelity test that validates or invalidates the yardstick.}
\label{fig:method}
\end{figure}

\looseness=-1 \sys treats code optimization under an expensive true objective as a search that must also decide \emph{what to optimize} and \emph{how to measure progress cheaply}.
It runs a fleet of coding agents, each co-discovering a target and a fast yardstick to climb, coordinated by a lightweight orchestrator that also runs \emph{fidelity tests} against the true evaluation (Figure~\ref{fig:method}).

\subsection{Problem: Optimization under expensive true evaluation}
\label{sec:formulation}

Let $\mathcal{X}$ be the space of program states, from a single file (e.g., kernel code) to an entire codebase, and let $x_0\in\mathcal{X}$ be the initial program state.
An \emph{edit} by an agent turns one state into another (e.g., updating files in a codebase).
The expensive \emph{true objective} $f:\mathcal{X}\to\mathbb{R}\cup\{\bot\}$ maps a state to a scalar score, where higher is better, or to $\bot$ (a failure value) if any correctness check fails.
We denote the cost of one call to $f$ by $c_f$ (e.g., the time it takes vLLM to serve a benchmark workload end to end).
Within a run, $N$ agents work under one shared wall-clock budget and evaluation resources.
For our experiments, $c_f$ ranges from minutes to hours (Table~\ref{tab:benchmarks}), so each agent gets only tens of true evaluations in total.
The goal of a run is to find a state that scores as high as possible under $f$.

\subsection{Co-discovery: Targets and their yardsticks}
\label{sec:codiscovery}

The building block of \sys is the worker agent, an off-the-shelf LLM coding agent, following recent agentic optimizers \citep{luo2026agentic,liao2026kernelevolve,deng2026perfagent,agarwal2026inductive}.
Each agent co-discovers a target and a yardstick for it and then climbs the yardstick to find solutions.
The optimization steps (profiling, choosing a target, building a yardstick, and making edits) are all performed by the agent, while the orchestrator adds measurement and coordination.

We denote a \emph{target} by $T$, which is a region of code (e.g., a single function in a kernel, or a module in a large codebase).
The set $\mathcal{X}_T\subseteq\mathcal{X}$ denotes the program states that can be reached by changing code only inside $T$.
Each agent comes up with its own target given the task and the initial program state $x_0$.
This is in contrast to AlphaEvolve~\citep{novikov2025alphaevolve}, where the user marks the code to be evolved for the system with explicit begin and end markers in the source.
In \sys, the agent also writes down a rough guess, $w_T\in[0,1]$, of what share of the true score the target region might be responsible for.
This is useful to know because, by Amdahl's law \citep{amdahl1967}, even if the target code region became infinitely fast, the overall score could improve by at most a factor of $1/(1-w_T)$ (i.e., a region that is responsible for 20\% of the score cannot buy more than 1.25$\times$).

We denote a \emph{yardstick} for $T$ by $g_T:\mathcal{X}_T\to\mathbb{R}\cup\{\bot\}$, a scorer that is much cheaper to run than the true evaluation, $c_g\ll c_f$.
The agent writes this as a script that returns a score and a correctness flag.
Each agent builds its own yardstick for its chosen target, and what it computes is up to the agent (e.g., a representative subset of cases, or a cost model).
The orchestrator measures how cheap a yardstick is as the ratio $r = c_f/c_g$ of the two costs, and passes it back to the agent, which decides whether to keep it.
The absolute value of a yardstick does not need to track $f$, only the order it puts edits in, since the yardstick is used only to decide which edits to pursue. For two edits $x,x'\in\mathcal{X}_T$, we want $\operatorname{sign}\!\left(g_T(x)-g_T(x')\right)=\operatorname{sign}\!\left(f(x)-f(x')\right)$.

By co-discovery we mean that the target and the yardstick are chosen together.
A concrete example of this co-discovery process is shared in Appendix~\ref{app:vllm}. When optimizing vLLM for performance, one agent targeted the scheduler's prefill batching and wrote a yardstick that was a simple model of decode-step time, which runs in seconds while the true evaluation takes 85 minutes.
Co-discovery is needed since a target is only valuable if a cheap yardstick that correlates with $f$ can be built for it, while a yardstick is only valuable for a target that can actually change $f$.

\subsection{Bottom-up search and coordination}
\label{sec:coordination}
Within a target, an agent runs two loops (Figure~\ref{fig:method}, right): a fast one, in which it edits and scores with its yardstick, and a slow one, in which a true evaluation confirms a gain in score that the yardstick reports.
In the fast loop, where it spends most of its time, it edits its target region, scores each edit with its yardstick $g_T$, keeps the edit if the score improves, and reverts it otherwise.

The search is bottom-up, in that each agent picks its own target without an a priori decomposition of the problem, so the orchestrator's role is coordination rather than planning.
Periodically, the orchestrator reads a short summary of every agent's progress and coordinates them through three actions.
It \emph{terminates} agents that are not making progress, \emph{spawns} new ones on promising or uncovered targets, and \emph{briefs} running ones with a short message about what the rest of the fleet has found.
A spawned agent (i) starts fresh, (ii) is seeded with the confirmed state of the best agent (a \emph{transfer}), or (iii) re-implements the confirmed changes of several agents in one place, keeping each only if a true evaluation confirms it (a \emph{combine}).
For example, over the course of the vLLM run, the orchestrator terminated one agent, briefed the other three to leave throughput and explore, and later spawned a fourth on a new target with a transferred solution.
A run ends when the budget is exhausted, a target score is reached, or the fleet empties.

\subsection{Fidelity testing: Validating yardsticks}
\label{sec:fidelity}

The fidelity test measures whether a yardstick ranks states in the same order as $f$, using Kendall's $\tau$ rank correlation \citep{kendall1938}, a standard tool in nonparametric statistics \citep{hollander2013nonparametric}.
We compute the correlation only on pairs of states whose true scores differ by more than the evaluation noise, which avoids giving credit for distinctions that the true evaluation itself cannot resolve.
A yardstick counts as correlated with $f$ when $\tau\ge 0.5$ over these pairs.

Since the search optimizes against its yardstick, it can be rewarded for edits that exploit the gap between the yardstick and $f$, a form of reward hacking \citep{gao2023scaling,skalse2022defining,karwowski2024goodhart}. Consequently, a yardstick that initially tracks $f$ may fail to do so for the specific edits targeted during search. %
For this reason, the orchestrator re-tests every yardstick after every true evaluation.
A new yardstick has not been tested yet, which does not stop the agent from climbing it.
Once enough states have been scored by both $g_T$ and $f$ and the correlation clears the bar, the yardstick is \emph{validated}. If the correlation later drops, it is \emph{invalidated}, and the agent either rebuilds it or moves on to a new target.
In the vLLM case, a yardstick built from a short load sweep was validated, later invalidated when higher load reversed its ranking, and validated again once rebuilt.
Appendix~\ref{app:fidelity} gives the noise estimation and thresholds.

\section{Results}\label{sec:results}

We detail the setup, then report benchmark results (\S\ref{sec:main_results}) and five real-world case studies (\S\ref{sec:deployment}).

\textbf{Benchmarks.} We evaluate \sys on tasks across two repository optimization benchmarks (\formulacode and \sweeff; \citealp{sehgal2026formulacode,ma2025swe}) and four GPU and TPU kernel suites (\solexec, \jax, \tpuinf, and \fa; \citealp{lin2026sol,tschand2026jaxbench,vllm_tpu_inference,chen2026avo}). Each task requires speeding up the source code while maintaining functional correctness. Unlike algorithmic benchmarks~\citep{novikov2025alphaevolve}, where evaluation finishes in seconds, a single true evaluation here can take minutes to hours (Table~\ref{tab:benchmarks}).

\textbf{Scoring.} All methods are evaluated under the same wall-clock time and number of agents. Repositories are scored by end-to-end execution speedup while checking for correctness. Kernels are scored by speedup over reference code or as a percentage of theoretical peak hardware throughput (speed-of-light or SOL score; \citealp{lin2026sol}). The reported score of a run is the median of several re-evaluations of its final candidate, chosen by lower confidence bound, with a Student-$t$ interval. The best score seen during the run is never reported, and a candidate that fails the correctness check is reported as no result. Search statistics and the analyses of \S\ref{sec:ablations} use a stratified sample of 20 \sweeff tasks and 50 \solexec kernels that covers every task type and evaluation time.

\textbf{Baselines.} We compare \sys against agentic (\claudecode; \citealp{claudecode2026}), evolutionary (\openevolve; \citealp{openevolve}; its agentic variant on the repository benchmarks), and domain-specific baselines, namely \avo~\citep{chen2026avo}, \kda~\citep{kda2025}, and a profiler-guided approach modeled on \perfagent~\citep{deng2026perfagent} (Appendix~\ref{app:baselines}). Unless noted, all runs execute for 8 hours on identical hardware with $N=4$ parallel agents driven by Claude Code v2.1 on claude-opus-5, with claude-sonnet-5 as the harness's small fast helper model, for \sys and for every Claude Code-based baseline. In \S\ref{sec:ablations}, we evaluate another backbone, Antigravity with Gemini 3.8 Flash~\citep{google2026gemini38flash}. Main results are single-run, and \S\ref{sec:ablations} reports variance across three runs on 50 \solexec kernels and 20 \sweeff tasks.

\subsection{Benchmark results}\label{sec:main_results}

\begin{table}[t]
\centering
\setlength{\tabcolsep}{3.0pt}
\vspace{-0.5\baselineskip}
\caption{Benchmark results under an 8-hour, four-agent budget (24 hours on the long \formulacode tasks, 10 hours on \tpuinf), higher is better. Numbers are speedups over the base code, except \fa (\% over Flash Attention 4), \jax (over XLA), \tpuinf (over tuned Pallas kernels), and \solexec, the mean \% of speed-of-light over 235 leaderboard kernels. Dashes mark benchmarks a method does not target.}
\label{tab:main}
\footnotesize
\setlength{\tabcolsep}{7pt}
\renewcommand{\arraystretch}{0.95}
\begin{tabular}{l cc cc cc}
\toprule
& \multicolumn{2}{c}{Repositories}
& \multicolumn{2}{c}{GPU Kernels}
& \multicolumn{2}{c}{TPU Kernels} \\
\cmidrule(lr){2-3}
\cmidrule(lr){4-5}
\cmidrule(lr){6-7}
Method & \formulacode & \sweeff & \solexec & \fa & \jax & \tpuinf \\
& ($\times$) & ($\times$) & (\% SOL) & (\%) & ($\times$) & ($\times$) \\
\midrule
Claude Code & 1.778 & 12.81 & 67.31 & 7.05 & \grounded{jax.claudecode}{3.82} & \grounded{tpuinf.claudecode}{2.38} \\
\openevolve & 1.161 & 3.76 & 52.36 & 1.48 & \grounded{jax.openevolve}{2.66} & \grounded{tpuinf.openevolve}{1.14} \\
\perfagent & 1.774 & 12.27 & 64.10 & 6.86 & \grounded{jax.perfagent}{4.08} & \grounded{tpuinf.perfagent}{2.14} \\
\kda & -- & -- & 70.22 & 8.93 & -- & -- \\
\midrule
\rowcolor{mainrow} \textbf{\systab (ours)} & \textbf{2.156} & \textbf{15.02} & \textbf{79.74} & \textbf{11.97} & \textbf{\grounded{jax.duodiscover}{4.62}} & \textbf{\grounded{tpuinf.duodiscover}{2.68}} \\
\bottomrule
\end{tabular}
\vspace{-0.6\baselineskip}
\end{table}

Under the same wall-clock budget, \sys scores above every baseline on every repository and kernel benchmark (Table~\ref{tab:main}). The rest of this section dives deeper into each of the benchmarks.

\textbf{FormulaCode.}
FormulaCode~\citep{sehgal2026formulacode} is a repository-level benchmark of real-world performance optimization tasks mined from merged, performance-improving pull requests across scientific Python projects. We evaluate on a 25-task subset of FormulaCode-V, the human-checked 108-task suite. Human expert PRs give a 1.048$\times$ geometric-mean speedup over the base code, and the goal is to exceed the human patch while maintaining functional correctness. Selected tasks have evaluation times from 7 minutes to 6.8 hours, and all baselines share the same budget, 8 hours for tasks under 2 hours and 24 hours for tasks over 2 hours.

On these tasks (Table~\ref{tab:main}), OpenEvolve barely clears the human bar at 1.161$\times$, while the agentic baselines average 1.78$\times$. \sys achieves 2.16$\times$, ahead of all baselines and the human PRs, a 21\% gain over the strongest agentic baseline. It also makes broader repository-level changes, modifying on average 22 files, 51 functions, and 1,800 lines per task, against more localized edits from PerfAgent (14 files, 800 lines) and OpenEvolve (2 files, 250 lines).

\textbf{\sweeff.} We evaluate \sys on the 100 \sweeff{}-Lite~\citep{ma2025swe} tasks, which require agents to speed up pull-request workloads in major Python libraries like NumPy and pandas without breaking tests.

Unlike FormulaCode, SWE-fficiency targets a single workload per task. \sys gives a 15.02$\times$ geometric-mean speedup, above the expert patches (6.90$\times$) and all baselines: Claude Code (12.81$\times$), \perfagent (12.27$\times$), and \openevolve (3.76$\times$). The tasks are difficult because locating bottlenecks in large codebases is hard,
and slow true evaluations (15 minutes on median, up to 93 minutes) strictly cap candidate testing within an 8-hour budget. \sys succeeds through search efficiency (Table~\ref{tab:search_summary}). On the 20-task sample, Claude Code crowds onto 2 to 4 targets per task and runs true evaluations on roughly 72 candidates to land 6 confirmed gains, while \sys isolates about 9 targets per task and builds a fast yardstick for each (5 of which pass the fidelity test). Because a yardstick runs in just 40 seconds on median, 22$\times$ faster than a true evaluation, agents test 358 candidates on it and send only 59 to true evaluation. This doubles confirmed gains to roughly 12 per task, for a 16\% higher geometric mean (14.93$\times$ against 12.85$\times$), and passes Claude Code's 8-hour score after 2.5 hours (median over 20 tasks).

\textbf{GPU and TPU kernels.} Finally, we extend our evaluation to kernel optimization tasks across GPUs and TPUs. Unlike multi-file repository tasks, a kernel is a single file, but it still has several stages, such as expert routing, gating, projections, and memory dispatch in a mixture-of-experts model, and a true evaluation runs the whole kernel every time (14 minutes on median, up to an hour on \solexec). \sys takes each stage as a target, builds a fast yardstick timing only that component, and hill-climbs on it before confirming gains on the complete kernel.
Similar to repository optimization on \sweeff, \sys searches more broadly with cheap yardsticks while using fewer true evaluations, leading to more confirmed gains and faster progress on SOL-Exec (Table~\ref{tab:search_summary}).

For GPU kernels, we evaluate on SOL-ExecBench (\solexec)~\citep{lin2026sol}, a recent GPU kernel benchmark of 235 kernels spanning emerging models and suites (e.g., FlashInfer-Bench; \citealp{xing2026flashinfer}). It scores implementations by the fraction of hardware speed of light (SOL) attained, and its public leaderboard enables direct comparison with external and proprietary submissions. At the time of submission\footnote{Our rankings are based on the SOL-ExecBench leaderboard state on September 10, 2026.}, \sys ranks first for 121 of the 235 kernels, second for 61, and third for 30 (Figure~\ref{fig:solexec_board}, Appendix~\ref{app:solexec}), with an average SOL score of 79.74\%, ahead of several proprietary entries. We also tested strong baselines, including \kda (70.22\% SOL, top-ranked in~\citet{flashinfer_contest_2026}).
We further evaluate on Flash Attention 4 (\fa), designed for Blackwell GPUs~\citep{zadouri2026flashattention} and not yet included in \solexec. Across 24 B200 configurations, \sys achieves an 11.97\% geometric-mean speedup over the \fa reference, ahead of \kda (8.93\%), NVIDIA's \avo~\citep{chen2026avo} (4.11\%), and cuDNN attention (1.30\%).

Moving beyond GPUs, we first evaluate on \jax, covering 17 priority JAX\slash TPU kernels. \sys achieves a \grounded{jax.duodiscover}{4.62}$\times$ geometric-mean speedup over the XLA baseline, ahead of \perfagent (\grounded{jax.perfagent}{4.08}$\times$), Claude Code (\grounded{jax.claudecode}{3.82}$\times$), and OpenEvolve (\grounded{jax.openevolve}{2.66}$\times$).
We further evaluate on the \tpuinf suite~\citep{vllm_tpu_inference}, which includes kernels from recent frontier models such as GLM-5.2 and Kimi K3, with reference implementations optimized by expert performance engineers. Speedups here are measured against those optimized production Pallas kernels. \sys attains a \grounded{tpuinf.duodiscover}{2.68}$\times$ speedup, against \grounded{tpuinf.claudecode}{2.38}$\times$ for Claude Code, \grounded{tpuinf.perfagent}{2.14}$\times$ for \perfagent, and \grounded{tpuinf.openevolve}{1.14}$\times$ for \openevolve.

\subsection{Real-world case studies}
\label{sec:deployment}

\begin{figure}[t]
\centering
\includegraphics{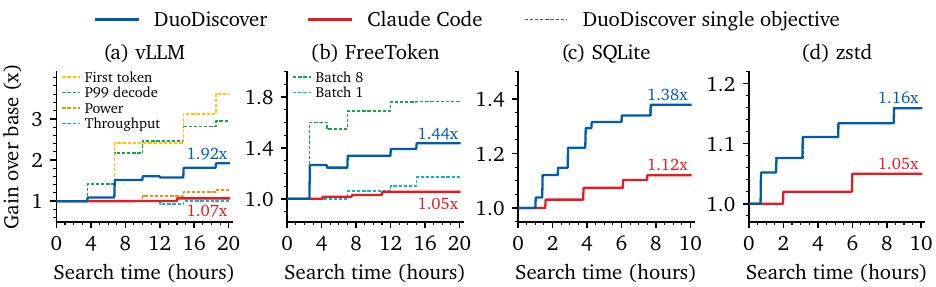}
\caption{Gain over the base code across search time. Thick lines: geometric mean over the objectives (\sys blue, Claude Code red). Dashed lines: individual \sys objectives.}
\label{fig:casestudies}
\end{figure}

In this section, we investigate whether \sys can find performance headroom in heavily optimized real systems and competitions like
vLLM~\citep[LLM serving]{kwon2023efficient},
FreeToken~\citep[on-device inference]{yang2026freetoken}, SQLite\footnote{\url{https://www.sqlite.org}} (embedded database), zstd\footnote{\url{https://github.com/facebook/zstd}} (compression), and data-prefetcher design for CPU microarchitectures~\citep{bera2026dpc4}, each of which has its own budget and baselines; 
all methods use four agents under the same budget, 20 hours for vLLM and FreeToken and 10 hours for SQLite and zstd. 
Every confirmed edit passes the system's or competition's full test suite and correctness checks. 
Maintainers of vLLM and FreeToken helped with initial reviews of our results, checking correctness and ruling out obvious reward hacking, and part of the FreeToken solution is already merged upstream. Search and solution details are in Appendix~\ref{app:deployment_details}.

\textbf{vLLM: Multi-Objective Serving under SLOs.} The task is to serve GLM-5.2 in NVFP4, NVIDIA's 4-bit floating-point format, on an 8$\times$B200 node under four competing service-level objectives (SLOs), namely 99th-percentile (p99) decode latency, first-token latency, GPU power, and throughput, scored by an 85-minute true evaluation. On this heavily tuned stack, four Claude Code agents improve neither p99 decode latency nor power and end at a 1.07$\times$ geometric mean after 12 true evaluations. \sys advances all four (Figure~\ref{fig:casestudies}(a)) and after 7 true evaluations lands a roughly 500-line, purely system-level solution at 1.92$\times$ geometric mean, with 2.94$\times$ lower p99 decode latency, 3.61$\times$ lower first-token latency, 21\% lower power, and unchanged throughput.

\textbf{FreeToken: On-Device LLM Inference.} FreeToken is an open-source engine that serves mixture-of-experts (MoE) models on a single GPU by keeping most weights in host memory~\citep{yang2026freetoken}. The task is to serve GLM-5.2-NVFP4 (a 753B-parameter MoE model with 433\,GB of weights) on a single 48\,GB L40S GPU, scored by a 70-minute true evaluation of decode throughput at batch sizes 1 and 8. Bound by CPU expert execution and PCIe transfers, Claude Code ends at 1.05$\times$ after 14 true evaluations. \sys finds complementary optimizations across the CPU expert path, GPU kernels, expert scheduling, and caching, and after 6 true evaluations ends at 1.17$\times$ at batch 1 and 1.76$\times$ at batch 8, a 1.44$\times$ geometric mean (Figure~\ref{fig:casestudies}(b)).\footnote{Part of this solution is merged into FreeToken: \url{https://github.com/FlashML-org/FreeToken/pull/619}.}

\textbf{SQLite and zstd.} Both are mature, heavily tuned C codebases, and zstd is the Zstandard lossless compression library. For SQLite, the task uses its own \texttt{speedtest1} workload, and each 10-minute true evaluation rebuilds the engine, runs the {13,249}-test suite twice, and checks hidden result hashes on unseen seeds and query shapes. \sys speeds up the bytecode engine, sorter, recursive-query path, and decimal conversion, ending at 1.38$\times$ against 1.12$\times$ for Claude Code, with 17 true evaluations against 45 (Figure~\ref{fig:casestudies}(c)). For zstd, each 15- to 25-minute true evaluation counts instructions across 16 workload configurations and requires byte-identical compressed output. \sys improves the optimal parser, match finder, and match compare, ending at a 1.16$\times$ instruction ratio against 1.05$\times$ for Claude Code, with 7 true evaluations against 21 (Figure~\ref{fig:casestudies}(d)).

\needspace{7\baselineskip}
\begin{wraptable}{r}{0.5\linewidth}
\centering\footnotesize
\setlength{\tabcolsep}{3.6pt}
\caption{Official DPC4 scores, geometric-mean performance gain over the reference core (610 traces for FullBW and LimitBW, 484 mixes for Multi-Core). BertiGO (the best human design) and ArchAgent v2 are as reported in the ArchAgent v2 paper~\citep{gonzalez2026archagent}.}
\label{tab:prefetcher}
\begin{tabular}{l cccc}
\toprule
& FullBW & LimitBW & Multi-Core & Overall \\
\midrule
BertiGO
    & +5.16\% & +2.57\% & +2.85\% & +3.52\% \\
ArchAgent v2
    & +5.61\% & +4.56\% & +1.33\% & +3.82\% \\
\midrule
\rowcolor{mainrow}
\textbf{\systab}
    & \textbf{+9.08\%}
    & \textbf{+5.93\%}
    & \textbf{+3.93\%}
    & \textbf{+6.31\%} \\
\bottomrule
\end{tabular}
\end{wraptable}
\textbf{Computer Architecture Design.} Following recent advances in autonomous chip design~\citep{goldie2024alphachip,openai-jalapeno,sankaralingam2026computer,team2026design},
we evaluate \sys{} on data-prefetcher optimization, a CPU microarchitecture design problem.
The task comes from DPC4~\citep{bera2026dpc4} (Table~\ref{tab:prefetcher}), a well-established competition in which prefetchers are scored on a common simulator, workload traces, and storage budget.
A single true evaluation takes up to \textbf{28 hours} running over 1{,}700 
simulations.
Performance is measured against the competition's reference CPU, which already runs the best prior published prefetchers.
Compared to ArchAgent v2~\citep{gonzalez2026archagent}, the previous best system on this task (built on AlphaEvolve), which evaluated over 12,000 prefetcher candidates over 40 days in the first of its five search stages,
\sys{} delivers a 6.3\% performance gain over the reference core (geometric mean across all three configurations), outperforming both it (+3.8\%) and the best human design (+3.5\%) while evaluating fewer candidates (8{,}000) over fewer days (16.7).
For scale, single-processor performance now improves by $\approx$3.5\%/year~\citep{hennessy2019new} and requires enormous investment\footnote{Designing a 5nm chip in 2020 cost over \$540M~\citep{chip-design-cost}.}. 
On multi-core workloads, \sys{} delivers a +3.9\% gain, ahead of the best human design (+2.9\%) and ArchAgent v2 (+1.3\%). To our knowledge, \sys{} is the first autonomous system to outperform the human best on these multi-core cases, a step toward autonomous chip design (Appendix~\ref{app:prefetcher}).

\begin{table}[t]
\centering\footnotesize
\setlength{\tabcolsep}{3.5pt}
\caption{Yardstick speed and fidelity across \sweeff tasks and \solexec kernels. Here \emph{Validated}: targets whose yardstick passes the fidelity test. \emph{Invalidated}: percentage invalidated later for drift. $\tau$: median rank correlation with the true evaluation.}
\label{tab:yardstick_quality}
\begin{tabular}{l c c c c c c}
\toprule
& True evaluation & Yardstick & Faster by & Validated & Invalidated & $\tau$ \\
\midrule
\sweeff tasks & 15 min & 40 s & 22$\times$ & 5.3 of 9.2 & 41\% & 0.66 \\
\solexec kernels & 14 min & 16 s & 53$\times$ & 11 of 15 & 45\% & 0.76 \\
\bottomrule
\end{tabular}
\end{table}

\begin{figure}[t]
\centering
\includegraphics{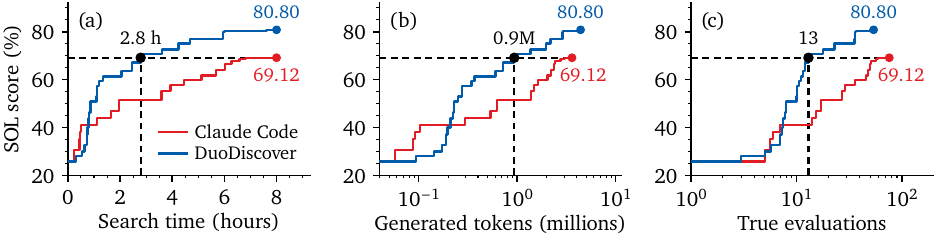}
\caption{Best-so-far SOL (mean) over 50 kernels vs.\ (a) time, (b) tokens, and (c) true evaluations. \sys attains Claude Code's score (69.12\%) in 2.8 hours, 0.93M tokens and 13 true evaluations.}
\vspace{-0.5\baselineskip}
\label{fig:main_tradeoff}
\end{figure}

\section{Analysis}\label{sec:ablations}
We evaluate \sys on 20 \sweeff tasks and 50 \solexec kernels, examining yardstick quality, search efficiency, variance, component ablations, and backbone generalization.

\textbf{Yardstick Quality.}
Yardsticks score candidate edits 22$\times$ faster on repositories (40 s vs.\ 15 min) and 53$\times$ faster on kernels (16 s vs.\ 14 min) than true evaluations (Table~\ref{tab:yardstick_quality}). Each measures only what its target affects, such as a shrunken input or a profiled hot loop, rather than the full workload. Yardsticks are validated on 5.3 of 9.2 repository targets and 11 of 15 kernel targets per task, and their rankings correlate closely with the true evaluation (median Kendall $\tau$ of 0.66 and 0.76). As code evolves, rank drift invalidates 41\% of repository yardsticks and 45\% of kernel yardsticks. This wastes budget but never counts a false gain, since every reported gain is confirmed by a true evaluation. Appendix~\ref{app:analysis} reports yardstick quality by kind (Table~\ref{tab:yardstick_kinds}) and per-candidate fidelity (Figure~\ref{fig:yardstick_quality}).

\textbf{Search Efficiency.}
\sys builds yardsticks in its first hour, then pulls ahead of Claude Code (Figure~\ref{fig:main_tradeoff}). On kernels, \sys equals Claude Code's final 69.12\% score in 2.8 hours using 74\% fewer tokens (0.93M vs.\ 3.6M) and 83\% fewer true evaluations (13 vs.\ 76), cutting run cost from \$309 to \$79.60 (Appendix~\ref{app:cost}). While faster loops let agents generate more tokens per hour, \sys turns them into real gains, ending at 80.80\% (4.4M tokens, 54 true evaluations) and leading on 41 of 50 kernels. On repositories, it passes Claude Code's score in a median of 2.5 hours.

\textbf{Variance across Discovery Runs.}
Across three seeds per task, \sys shows low run-to-run variance with final scores staying within $\pm$1.12\% of its per-kernel mean and $\pm$1.8\% of its per-task mean, against $\pm$3.27\% and $\pm$5.21\% for Claude Code.
Most gains reproduce across seeds, with 78\% on kernels and 81\% on repositories.

\textbf{Component Ablations.}
We add components incrementally (Table~\ref{tab:ablation}) and remove orchestrator actions one by one (Table~\ref{tab:orch}). Target selection alone changes little,
as true evaluations still bottleneck candidate testing. Untested yardsticks hurt
because rank drift misguides search. The fidelity test recovers this loss (13.25$\times$, 72.57\%), co-discovery adds more (13.76$\times$, 74.43\%), and the orchestrator adds +1.17$\times$ and +6.37 SOL points, the largest single step on kernels, to the full \sys (14.93$\times$, 80.80\%). Removing any single action costs 0.63 to 1.01$\times$ in repository speedup and 2.3 to 4.4 kernel points, hitting transfer hardest on repositories and candidate combination on kernels. The removal losses sum to more than the orchestrator's net gain, so its actions depend on one another.

\begin{table}[t]
\centering\footnotesize
\captionsetup{justification=raggedright} %
\setlength{\tabcolsep}{3pt}
\sbox0{\begin{tabular}{@{}l c c@{}}
\toprule
& \sweeff & \solexec \\
& ($\times$) & (\% SOL) \\
\midrule
Claude Code ($N=4$) & 12.85 & 69.12 \\
\quad {}+ target & 12.71 & 69.68 \\
\quad {}+ yardstick (no fidelity test) & 10.33 & 66.91 \\
\quad {}+ fidelity test & 13.25 & 72.57 \\
\quad {}+ co-discovery & 13.76 & 74.43 \\
\rowcolor{mainrow} \quad {}+ orchestrator (\systab) & \textbf{14.93} & \textbf{80.80} \\
\bottomrule
\end{tabular}}%
\sbox2{\begin{tabular}{@{}l c c@{}}
\toprule
& \sweeff & \solexec \\
& ($\times$) & (\% SOL) \\
\midrule
\rowcolor{mainrow} \systab~(full) & \textbf{14.93} & \textbf{80.80} \\
\quad $-$ spawn and terminate & 14.10 & 77.93 \\
\quad $-$ allocate by progress & 14.30 & 78.23 \\
\quad $-$ brief & 14.28 & 78.49 \\
\quad $-$ transfer & 13.92 & 77.46 \\
\quad $-$ combine & 14.03 & 76.38 \\
\bottomrule
\end{tabular}}%
\begin{minipage}[t]{\wd0}
\captionof{table}{Adding one component at a time, up to the full \sys (shaded).}
\label{tab:ablation}
\usebox0
\end{minipage}\hspace{2.5em}%
\begin{minipage}[t]{\wd2}
\captionof{table}{Removing one mechanism at a time from the full \sys (shaded).}
\label{tab:orch}
\usebox2
\end{minipage}
\end{table}

\needspace{15\baselineskip}
\begin{wraptable}{r}{0.5\linewidth}
\centering\footnotesize
\setlength{\tabcolsep}{4pt}
\caption{Generalization across coding-agent backbones under the 8-hour, 4-agent budget.}
\label{tab:backbone}
\begin{tabular}{l cc}
\toprule
& \sweeff & \solexec \\
& ($\times$) & (\% SOL) \\
\midrule
Claude Code/Opus 5 & 12.85 & 69.12 \\
\rowcolor{mainrow} +\systab & \textbf{14.93} & \textbf{80.80} \\
\cmidrule(lr){1-3}
Antigravity/Gemini 3.8 Flash & 9.72 & 58.79 \\
\rowcolor{mainrow} +\systab & \textbf{12.66} & \textbf{73.26} \\
\bottomrule
\end{tabular}
\end{wraptable}
\textbf{Backbone Coding-Agent Generalization.}
We also run \sys on
Antigravity\slash Gemini 3.8 Flash, a different coding-agent backbone, under the same 8-hour, 4-agent budget
(Table~\ref{tab:backbone}). With Antigravity, \sys raises
repository speedup from 9.72$\times$ to 12.66$\times$ and kernel
SOL score from 58.79\% to 73.26\%, corresponding to
1.30$\times$ and 1.25$\times$ gains, larger than the 1.16$\times$ and
1.17$\times$ gains with Claude Code\slash Opus 5. \sys largely closes
the gap between the two agents. With \sys, Antigravity\slash Gemini 3.8 Flash
nearly matches Claude Code\slash Opus 5 on repositories
(12.66$\times$ vs.\ 12.85$\times$) and surpasses it on kernels
(73.26\% vs.\ 69.12\%). As expected for the faster model,
Gemini 3.8 Flash generates 32\% more tokens, but its 6.7$\times$ lower per-token API cost makes the
overall run 5.1$\times$ cheaper than Claude Code\slash Opus 5
(Appendix~\ref{app:cost}).

The appendix extends this analysis. Appendix~\ref{app:ablations} scales the fleet from 1 to 16 agents and compares per-target yardsticks with a single task-wide evaluator, and Appendix~\ref{app:analysis} reports yardstick quality by kind, search behavior, the cost of a run, examples of discovered targets and yardsticks, and robustness to missing or unreliable yardsticks and noisy evaluations.

\section{Related work}
\label{sec:related_work}

In this section, we describe three lines of related research: LLM-driven coding and
discovery agents, LLM-generated tests and evaluators, and multi-fidelity optimization methods.

\textbf{Coding and discovery agents.}
LLMs are increasingly used as operators for search and optimization, transforming automated code generation into an engine for scientific and algorithmic discovery. FunSearch \citep{romera2024funsearch} was among the first methods of this kind and found new best-known constructions for combinatorics problems.
AlphaEvolve \citep{novikov2025alphaevolve} scaled this into an autonomous pipeline that evolves a population of programs from programmatic feedback.
Subsequent work has extended this evolutionary approach along several axes: improving efficiency \citep{lange2026shinkaevolve, jiang2026deltaevolve}, dynamically adapting mutation strategies \citep{cemri2026adaevolve,liu2026evox}, optimizing the model weights with test-time reinforcement learning \citep{wang2026thetaevolve, yuksekgonul2026learning}, and overcoming local minima \citep{yan2026pacevolve}.
Discovery agents have been proposed for a variety of tasks, including repository-level runtime optimization \citep{deng2026perfagent,lin2025eco}, low-level kernel acceleration \citep{kda2025, chen2026avo,wang2026maxkernel, luo2026agentic, dai2026cuda, liao2026kernelevolve,wei2025astra}, and computer architecture discovery \citep{gupta2026archagent, gonzalez2026archagent, blasberg2026agentic}.
For some tasks, the cost of evaluations can be a major bottleneck for optimization. To mitigate this bottleneck, SIFT \citep{fu2026self} uses the expensive evaluations only for the most promising solutions.
While prior methods like AlphaEvolve allow users to specify cheaper checks ahead of the full evaluation, \sys is the first discovery agent that finds both the solution and cheap evaluators which are validated or invalidated during the discovery process.

\textbf{LLM-generated tests and evaluators.}
Prior work has also used LLMs to synthesize test cases and evaluators to guide code search.
Generated unit tests are used to rank candidate implementations \citep{chen2023codet} and to drive iterative refinement \citep{shinn2023reflexion, chen2024teaching, huang2023agentcoder, ruan2025specrover, ehrlich2025codemonkeys}.
Alternatively, ExecCritic \citep{tao2026execcritic} freezes generated tests before changing the implementation so unverified checks cannot misguide revisions \citep{chen2026rethinking}.
While these works use LLMs and coding agents to construct software correctness tests, \sys extends this idea beyond correctness testing to discover cheap, target-specific yardsticks for fast optimization.

\textbf{Multi-fidelity optimization (MFO).}
MFO methods balance cheap, approximate evaluations and expensive, accurate ones by spending most of the budget on a low-cost proxy and querying the expensive objective sparingly \citep{li2026multi}.
There are two common approaches. The first is to fit a cheap \emph{surrogate model} that predicts the objective and is corrected as real measurements arrive \citep{kennedy2000predicting,forrester2007multi,kandasamy2016gaussian,chen2018tvm,fernandez2019issues}. The second is \emph{adaptive resource allocation}, which evaluates many candidates cheaply, then progressively concentrates budget on the promising ones and discards the rest \citep{li2018hyperband,falkner2018bohb}.
Uniquely, our method uses agents to synthesize cheap proxy evaluators and validate or invalidate them during search.

\section{Conclusion}
\looseness=-1
Scaling autonomous discovery requires not only finding better solutions, but also discovering how to evaluate them efficiently.
In this work, we have argued that search in this regime must discover both \textit{what to optimize (target)} and \textit{a cheaper way to measure progress on it (yardstick)}. We present \sys, a bottom-up search system that builds on this idea for evaluation-bottlenecked settings. Across two repository benchmarks, four kernel suites, and five real-world case studies, 
\sys demonstrates autonomous optimization across the computing stack, from software to chip architecture.
More broadly, \textit{co-discovery}, enabling agents to find their own targets and yardsticks, opens a path toward scaling autonomous discovery across science and engineering.

\textbf{Limitations.} While \sys{} accelerates optimization across diverse domains, it has two primary limitations. First, when no faithful yardstick can be constructed for a target, \sys{} falls back to standard search over the expensive true objective. Second, constructing yardsticks and running initial fidelity tests incurs an upfront warmup overhead, making \sys{} less advantageous under extremely short evaluation budgets.

\section*{Acknowledgments}
We thank our colleagues at Google Systems Research Group, Cloud AI Research, and across the broader AI and Infrastructure organization for valuable feedback and discussions.

\setlength{\bibsep}{2.4pt} %
\bibliography{references}
\clearpage %
\noindent{\headingfont\fontsize{15}{17}\selectfont Appendix}\par\vspace{-1ex}
\addcontentsline{toc}{section}{Appendix}

\appendix

\section{Method details}\label{app:method}
This appendix expands \S\ref{sec:method}, and the notation dictionary is given in Table~\ref{tab:notation}.
A \emph{target} is what an agent optimizes, a region of the code, and a \emph{yardstick} is the cheap scorer the agent builds for it (\S\ref{sec:codiscovery}); choosing the two together is \emph{co-discovery}.
The \emph{fidelity test} is the orchestrator's rank-correlation test of a yardstick against $f$, repeated after every true evaluation. A yardstick that has passed it is \emph{validated}, and one that fails it after having passed is \emph{invalidated} (\S\ref{sec:fidelity}).
The \emph{orchestrator} is an LLM, with supporting code, that coordinates the agents and runs the fidelity test (\S\ref{sec:method}).

\begin{table}[!h]  %
\centering\footnotesize
\caption{Notation of \S\ref{sec:method}, in the order it is introduced.}
\label{tab:notation}
\begin{tabular}{l p{8.1cm} l}
\toprule
Symbol & Meaning & Defined in \\
\midrule
$x$, $\mathcal{X}$ & a program state (a version of the code); the set of all states & \S\ref{sec:formulation} \\
$x_0$ & the initial program & \S\ref{sec:formulation} \\
$f$ & the true evaluation, $f:\mathcal{X}\to\mathbb{R}\cup\{\bot\}$, higher is better; $\bot$ means the state fails the task's correctness check & \S\ref{sec:formulation} \\
$c_f$ & the cost of one call to $f$, measured as its running time & \S\ref{sec:formulation} \\
$N$ & the number of agents in a run & \S\ref{sec:formulation} \\
$T$, $\mathcal{X}_T$ & a target (a region of the code); the states reachable by editing only inside $T$ & \S\ref{sec:codiscovery} \\
$w_T$ & the agent's rough guess of the share of the true score that $T$ accounts for & \S\ref{sec:codiscovery} \\
$g_T$ & the yardstick for $T$, $g_T:\mathcal{X}_T\to\mathbb{R}\cup\{\bot\}$ & \S\ref{sec:codiscovery} \\
$c_g$, $r=c_f/c_g$ & the cost of one yardstick run; how many yardstick runs fit in one true evaluation & \S\ref{sec:codiscovery} \\
$\tau$ & Kendall's rank correlation between $g_T$ and $f$ over the resolved pairs (Eq.~\ref{eq:tau}) & \S\ref{sec:fidelity} \\
\bottomrule
\end{tabular}
\end{table}

\subsection{Fidelity test details}\label{app:fidelity}
This section gives the details of the fidelity test of \S\ref{sec:fidelity}: how the noise floor is measured, which statistic and significance test we use, and when a yardstick is validated or invalidated and what follows.

Let $S=\{x_1,\ldots,x_n\}$ be the states scored by both $f$ and the current yardstick, pooled across all agents that share the yardstick, and let $\hat{\sigma}$ be the relative evaluation noise, estimated from repeated true evaluations of the same state.
Only pairs that the true evaluation itself can resolve are counted:
\begin{equation}\label{eq:tau}
P_{\hat{\sigma}}(S)
=
\left\{
(i,j):
|f(x_i)-f(x_j)|
>
2\hat{\sigma}\bar{f}_{ij}
\right\},
\qquad
\tau(g_T,f)
=
\frac{n_c-n_d}{n_c+n_d},
\end{equation}
where $\bar{f}_{ij}$ is the mean score of the pair, and $n_c$ and $n_d$ count the pairs in $P_{\hat{\sigma}}(S)$ that $g_T$ orders the same way as $f$ and the opposite way, respectively (pairs tied under the yardstick are excluded).
Two states within noise of each other can therefore neither add to nor subtract from the correlation.
Throughout this appendix, $\tau$ means this restricted statistic.

\textbf{Noise floor.}
The relative evaluation noise $\hat\sigma$ is measured from the run itself.
Whenever the same snapshot (identified by a content hash of its source files) has been evaluated more than once, the residuals of $\log f$ across those evaluations are pooled over all such snapshots, and $\hat\sigma$ is the 95\% upper confidence bound on their standard deviation, so that a lucky low sample cannot understate the noise.
Until at least one snapshot has been evaluated twice, the noise is unmeasured and no yardstick is validated. A minimum separation of 1\% applies in any case, so that sub-percent jitter never counts as a resolved difference.
A pair of states is resolved when its true scores differ by more than $2\hat\sigma$ times their mean (Eq.~\ref{eq:tau}).

\textbf{Statistic and significance test.}
Rank correlation between a proxy and the objective is a commonly used test \citep{abdelfattah2021zerocost,white2021powerful,akhauri2022eznas,shen2023proxybo}.
All runs in this paper use Kendall's $\tau$ \citep{kendall1938} over the resolved pairs, with yardstick ties excluded, and a one-sided permutation test in each direction.
The test permutes the yardstick scores across states and recomputes the restricted $\tau$ under each permutation; the $p$-value is the share of permutations at or beyond the observed value.

\textbf{When a yardstick is validated or invalidated.}
A yardstick is \emph{validated} when the noise has been measured, at least $m=6$ distinct states resolve beyond it, $\tau\ge\tau_{\min}=0.5$, and the permutation test rejects independence at level $\alpha=0.05$.
Until then the yardstick is simply untested, and the agent is told what evidence would settle the question, typically more confirmed states that are far apart.
A yardstick is \emph{invalidated} when it has that much evidence yet $\tau<\tau_{\min}$. A significantly negative $\tau$ also forces a switch to a different kind of yardstick at once.
After an invalidation, the orchestrator stops spending true evaluations on confirming states climbed under that yardstick, and the agent either rebuilds the yardstick or moves on to a new target. The agent sees the measured $\tau$ and the ratio $r$ whenever it is launched, and it may rebuild on its own before the test has reached a conclusion.
If no yardstick for the target validates, the agent scores candidates with $f$ alone.

\section{Benchmark and baseline details}
This appendix expands the setup of \S\ref{sec:results}. It lists the benchmarks (Table~\ref{tab:benchmarks}) and how each baseline was run (Table~\ref{tab:baselines}).

\begin{table}[t]
\centering\footnotesize
\setlength{\tabcolsep}{3pt}
\caption{The benchmarks, with their task type, score, comparison baseline, task count $n$, budget, and true-evaluation time. On each benchmark, every method we run shares the same hardware, budget, agent count (four), and concurrency limit (four concurrent true evaluations on the benchmarks, one on the case studies). A true evaluation is one sequential run of the full workload, so adding machines cannot shorten it. The data-prefetcher study is the exception. It uses three agents, runs its traces in parallel, and takes its baselines as reported rather than rerun.}
\label{tab:benchmarks}
\begin{tabularx}{\linewidth}{l l l >{\raggedright\arraybackslash}X c c c}
\toprule
Benchmark & Task & Score & Compares against & $n$ & Budget & Eval time \\
\midrule
FormulaCode   & repo runtime & speedup ($\times$) & base code & 25 & 8 h / 24 h & 7 min - 6.8 h \\
SWE-fficiency & repo runtime & speedup ($\times$) & base code & 100 & 8 h & 15 min (3 to 93) \\
SOL-Exec      & GPU kernel   & \% of SOL          & hardware peak & 235 & 8 h & 14 min (2 to 60) \\
\jax           & kernel       & speedup ($\times$) & XLA baseline & \grounded{jaxbench.task_count}{17} & \grounded{jaxbench.budget_h}{8} h & \grounded{jaxbench.eval_time_form}{17 min (7 to 32)} \\
\tpuinf & kernel       & speedup ($\times$) & optimized Pallas kernel & \grounded{tpuinf.task_count}{7} & \grounded{tpuinf.budget_h}{10} h & \grounded{tpuinf.eval_time_form}{3 min (1 to 8)} \\
FA4           & kernel       & \% over FA4        & Flash Attention 4 & 1 & 8 h & 4 min \\
vLLM & open-source & geomean gain ($\times$) & tuned, unmodified vLLM & 1 & 20 h & about 85 min \\
FreeToken & open-source & geomean gain ($\times$) & unmodified FreeToken & 1 & 20 h & about 70 min \\
SQLite        & open-source & speedup ($\times$) & base code  & 1 & 10 h & about 10 min \\
zstd          & open-source & instruction ratio ($\times$) & base code  & 1 & 10 h & 15 to 25 min \\
Data prefetcher & hardware design & speedup ($\times$) & reference core & 1 & 16.7 d & up to 28 h \\
\bottomrule
\end{tabularx}
\end{table}

\subsection{Benchmarks}
\label{app:setup}
We report every benchmark in Table~\ref{tab:benchmarks}: its score, reference, task count, budget, and true-evaluation time. On \solexec, the unmodified reference code averages 26\% of SOL over the 235 kernels.

\textbf{Hardware.} \solexec and \fa kernels run and are evaluated on NVIDIA B200 GPUs, one GPU per kernel. \tpuinf and \jax kernels run on TPU v6e. Repository tasks are timed one per machine, on 16-vCPU, 32~GB nodes for \sweeff and 2-vCPU, 4~GB nodes for \formulacode. The case studies use an 8$\times$B200 node for vLLM, one 48\,GB L40S, with the model's 433\,GB of weights in host memory, for FreeToken, and a 44-vCPU node for SQLite and zstd (Appendix~\ref{app:deployment_details}). The data-prefetcher study runs on the ChampSim simulator~\citep{gober2022championship} on a 3000-vCPU cluster in three configurations, one core at full bandwidth, one bandwidth-limited, and four cores sharing memory. Every method we run on a benchmark uses the same hardware, and the true-evaluation times in Table~\ref{tab:benchmarks} are measured on it.

\textbf{Budget.} On the benchmarks every method gets 8 hours (24 hours on the \formulacode tasks whose true evaluation exceeds 2 hours, and 10 hours on \tpuinf), four agents, and at most four concurrent true evaluations. The other four case studies run true evaluations one at a time, since each uses a single machine (the 8-GPU node for vLLM, the L40S for FreeToken, and one node for SQLite and zstd). The data-prefetcher study instead runs true evaluations in parallel on a cluster, each fanning its 610 traces and 484 mixes across cores, so we report both its wall-clock time and its total CPU-hours (Appendix~\ref{app:prefetcher}). The orchestrator also manages the agent and evaluation budget, directing more agents and true evaluations toward the targets that are making progress; the ``allocate by progress'' ablation in Table~\ref{tab:orch} removes this.

\subsection{Baselines}
\label{app:baselines}
Table~\ref{tab:baselines} details each evaluated baseline alongside its execution setup. Every baseline executed within our benchmark suite shares \sys's hardware setup, budget (8 hours on the benchmarks and 20 or 10 hours on the case studies, with per-benchmark exceptions in Table~\ref{tab:benchmarks}), agent count, base code, task prompts, and true-evaluation procedures. Every method we run is evaluated on the full set of each benchmark it targets, all 100 \sweeff tasks or all 235 \solexec kernels, except Antigravity, which runs on the 20-task and 50-kernel sample. The \fa numbers for \avo and cuDNN are adopted directly from the \avo paper.

\begin{table}[t]
\centering\footnotesize
\setlength{\tabcolsep}{4pt}
\caption{Summary of baseline methods, benchmark coverage, and execution configurations.}
\label{tab:baselines}
\begin{tabularx}{\linewidth}{l l X}
\toprule
Baseline & Benchmarks & Setup \\
\midrule
\claudecode & all & Four independent agents on Opus 5, same prompt as \sys, each on its own copy of the code. Every candidate scored by a true evaluation. Best of four reported. \\ \specialrule{0.2pt}{2pt}{2pt}
\openevolve & all & On \formulacode and \sweeff we run \openevolveagentic, with the mutation operator replaced by a coding agent so it can edit whole codebases. On the kernel suites, unmodified \openevolve. Both run on Opus 5 under the same budget and true-evaluation setup as the other baselines. True evaluation as fitness, best of four independent runs reported. \\ \specialrule{0.2pt}{2pt}{2pt}
\perfagent & all & Our reimplementation on Opus 5, four agents. Each profiles with the task's profiler, edits the hottest region, re-profiles. Every candidate scored by a true evaluation. \\ \specialrule{0.2pt}{2pt}{2pt}
\kda & \solexec, \fa & Run by us, 8-hour budget, same hardware. Same model and agents as \sys, run as KDA kernel agents. \\ \specialrule{0.2pt}{2pt}{2pt}
\avo & \fa & As reported in the \avo paper. \\ \specialrule{0.2pt}{2pt}{2pt}
cuDNN & \fa & NVIDIA's attention kernels, no search. As reported in the \avo paper. \\ \specialrule{0.2pt}{2pt}{2pt}
Antigravity & \sweeff, \solexec & Antigravity harness on Gemini 3.8 Flash, run as \claudecode above, on the 20-task and 50-kernel sample (\S\ref{sec:ablations}). \\ \specialrule{0.2pt}{2pt}{2pt}
ArchAgent v2 & data prefetcher & AlphaEvolve-based system; its first of five search stages alone ran over 40 days and evaluated 12{,}000+ candidates. Numbers as reported~\citep{gonzalez2026archagent}. \\ \specialrule{0.2pt}{2pt}{2pt}
BertiGO & data prefetcher & Best human-designed prefetcher, no search~\citep{singh2026pushing}. Numbers as reported in the ArchAgent v2 paper~\citep{gonzalez2026archagent}. \\
\bottomrule
\end{tabularx}
\end{table}

\section{Extended main results}\label{app:per_task_results}
This appendix expands \S\ref{sec:main_results}, one subsection per benchmark in the order of Table~\ref{tab:main}.

\subsection{\formulacode}\label{app:formulacode}
Per-task scores for every method on the 25 \formulacode tasks can be seen in Table~\ref{tab:fc-speedup-comparison}. Each task provides a snapshot of the repository just before the human expert PR, the original issue and discussion context, correctness tests, and a suite of community-maintained Airspeed Velocity (ASV) performance benchmarks. FormulaCode exposes an average of 264 performance workloads per task and scores a solution by its geometric-mean speedup across these workloads. \sys beats the other baselines in 16 out of 25 tasks, with 21\% higher mean speedup across the selected tasks.

\begin{table}[t]
\centering
\footnotesize
\setlength{\tabcolsep}{8pt}
\renewcommand{\arraystretch}{1.0}
\caption{Speedup comparison across 25 tasks for \formulacode. Entries report geometric mean speedups (higher is better). The best result in each row is bold. Mean Speedup is the arithmetic mean of the task scores, and Wins counts first-place finishes.}
\label{tab:fc-speedup-comparison}
\begin{tabular}{@{}lcccc@{}}
\toprule
Task & \openevolve & \claudecode & \perfagent & \systab \\
\midrule
\texttt{TileDB-Py\_2206} & 1.175 & 1.627 & 1.461 & \textbf{1.730} \\
\texttt{uxarray\_1249} & 1.116 & 3.207 & 4.093 & \textbf{6.076} \\
\texttt{networkx\_6879} & 1.430 & 3.349 & 2.310 & \textbf{3.554} \\
\texttt{networkx\_7181} & 1.056 & 1.652 & 1.566 & \textbf{2.627} \\
\texttt{networkx\_7249} & 1.420 & 2.137 & 1.984 & \textbf{3.666} \\
\texttt{networkx\_7595} & 1.015 & 1.607 & 2.237 & \textbf{2.616} \\
\texttt{networkx\_7971} & 1.008 & \textbf{2.557} & 1.647 & 2.182 \\
\texttt{networkx\_8077} & 1.098 & 1.815 & 1.990 & \textbf{2.218} \\
\texttt{networkx\_8206} & 1.001 & 1.341 & \textbf{1.404} & 1.305 \\
\texttt{PyBOP\_301} & 1.707 & \textbf{2.748} & 2.557 & 2.108 \\
\texttt{bottleneck\_329} & 1.038 & \textbf{1.478} & 1.385 & 1.193 \\
\texttt{pymc\_6711} & 1.742 & 2.682 & 2.913 & \textbf{3.947} \\
\texttt{pymc\_7578} & 1.585 & \textbf{2.527} & 2.334 & 2.399 \\
\texttt{shapely\_2283} & 1.068 & 2.034 & 1.078 & \textbf{2.097} \\
\texttt{flox\_300} & 1.049 & 1.145 & \textbf{3.292} & 2.064 \\
\texttt{flox\_360} & 1.075 & \textbf{2.416} & 1.004 & 2.344 \\
\texttt{flox\_415} & 1.052 & 1.005 & 2.067 & \textbf{2.281} \\
\texttt{xdsl\_4909} & 1.357 & 1.826 & 1.776 & \textbf{1.971} \\
\texttt{pandas\_51344} & 1.015 & 1.023 & 1.031 & \textbf{1.039} \\
\texttt{pandas\_53397} & 1.006 & 1.045 & 1.036 & \textbf{1.089} \\
\texttt{pandas\_55104} & 1.016 & 0.993 & \textbf{1.068} & 1.041 \\
\texttt{pandas\_57651} & 1.002 & 1.020 & 1.027 & \textbf{1.061} \\
\texttt{pandas\_59647} & 1.001 & 1.024 & 1.029 & \textbf{1.131} \\
\texttt{pandas\_60461} & 1.004 & \textbf{1.132} & 1.028 & 1.052 \\
\texttt{pandas\_62623} & 0.999 & 1.071 & 1.028 & \textbf{1.100} \\
\midrule
\textbf{Mean Speedup} & 1.161 & 1.778 & 1.774 & \textbf{2.156} \\
\textbf{Wins (out of 25)} & 0 & 6 & 3 & \textbf{16} \\
\bottomrule
\end{tabular}
\setlength{\abovecaptionskip}{5pt}
\setlength{\belowcaptionskip}{0pt}
\end{table}

\subsection{\sweeff}\label{app:sweff}
A true evaluation runs the repository's own tests that cover the changed code and times the task's workload. A candidate that fails a test scores base speed, and a score of 1.00 means no gain over the base code. \sys's final patch passes the tests on all 100 tasks.

\subsection{\solexec}\label{app:solexec}
\begin{figure}[t]
\centering
\includegraphics{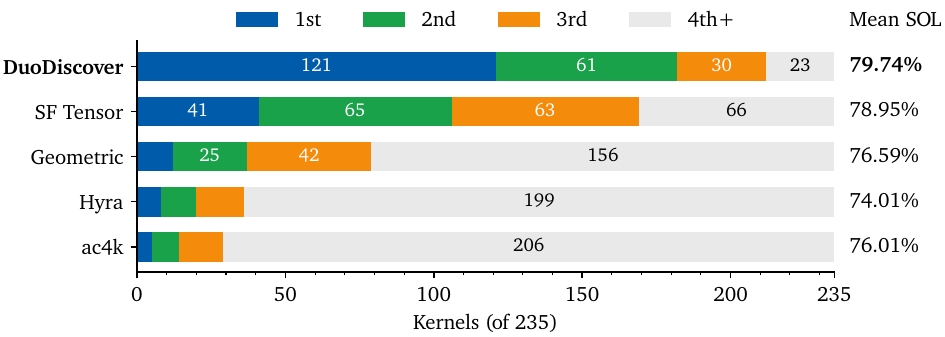}
\caption{Per-kernel finishing position of the top five \solexec entries, with mean SOL at the bar.}
\label{fig:solexec_board}
\end{figure}

On September 10, 2026, \sys held first place on the public \solexec leaderboard (Figure~\ref{fig:solexec_board}), with a mean SOL score of 79.74\% over all 235 kernels against 78.95\% for the second entry. The entry is our single 8-hour, four-agent run per kernel from Table~\ref{tab:main}, submitted once. Every method's SOL scores are computed by the same public leaderboard. The lead comes from breadth rather than a few kernels (Figure~\ref{fig:solexec_board}). \sys ranks in the top three on 212 of the 235 kernels, the second entry on 169, and it holds first place on 121 kernels against the runner-up's 41. Across the leaderboard's four collections it is first on L1, L2 and FlashInfer and second on Quant.

\subsection{\fa}\label{app:fa4}
\begin{figure}[t]
    \centering
    \includegraphics{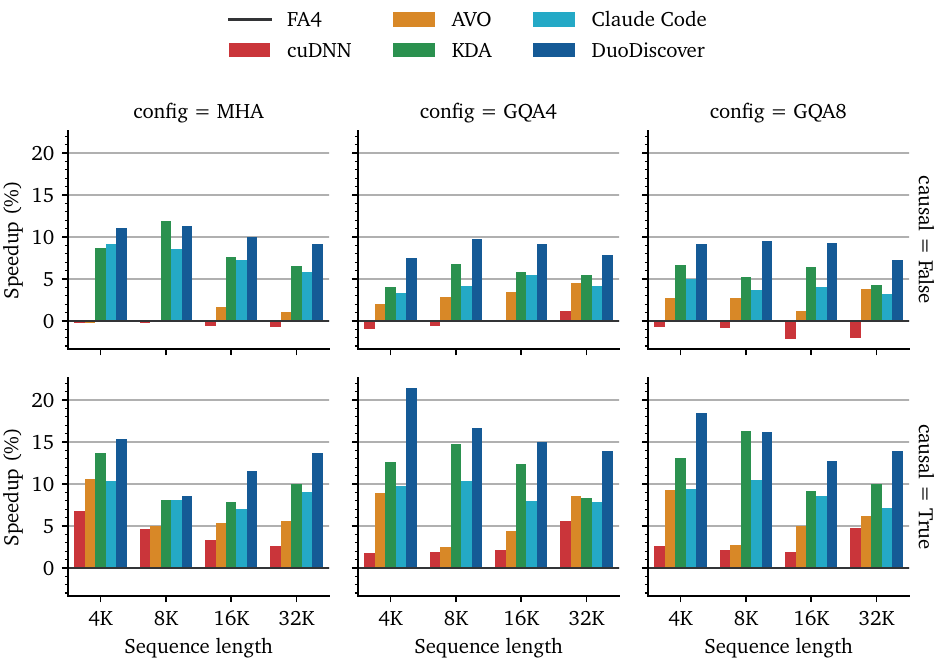}
    \caption{Speedup over Flash Attention 4 on each of the 24 \fa configurations.}
    \label{fig:fa4_bench}
\end{figure}

We compare every method on the 24 \fa configurations, each scored as a speedup over Flash Attention 4, and present the geometric mean in Table~\ref{tab:fa4} and the per-configuration speedups in Figure~\ref{fig:fa4_bench}. \sys is 11.97\% faster than Flash Attention 4, ahead of \kda at 8.93\%, Claude Code at 7.05\%, \perfagent at 6.86\%, \openevolve at 1.48\%, and the two entries reported in the \avo paper, \avo at 4.11\% and cuDNN at 1.30\%.

\begin{table}[t]
\centering\footnotesize
    \caption{Geometric-mean speedup over Flash Attention 4 across the 24 \fa configurations. cuDNN and \avo are as reported in the \avo paper~\citep{chen2026avo}. The rest are our runs.}
    \label{tab:fa4}
    \begin{tabular}{lc}
        \toprule
        Method & Speedup over FA4 (\%) \\
        \midrule
        FA4 & 0.00 \\
        cuDNN & 1.30 \\
        AVO & 4.11 \\
        KDA & 8.93 \\
        Claude Code & 7.05 \\
        \perfagent & 6.86 \\
        \openevolve & 1.48 \\
        \textbf{\systab (ours)} & \textbf{11.97} \\
        \bottomrule
    \end{tabular}
\end{table}

\subsection{\jax}\label{app:jax}
We evaluate every method on all 17 priority kernels of \jax~\citep{tschand2026jaxbench}, a suite of production JAX/TPU operators drawn from models such as Llama-3.1, DeepSeek-V3, Mixtral-8x7B, Mamba-2, and RetNet (attention variants, mixture-of-experts, general matrix multiplication (GEMM), normalization, and more). Each kernel is evaluated on 13 input shapes taken from production use, with random seeds that the agent never sees. The correctness gate adopts \solexec's setup~\citep{lin2026sol}: a candidate passes only if at least 98\% of its output elements lie within a tolerance, set per data type, of the XLA baseline's output. Its score is the geometric-mean speedup over that baseline across the 13 shapes (XLA $=1.00\times$). The per-method geometric means are in Table~\ref{tab:main}.

\subsection{\tpuinf}\label{app:tpu}
\begin{figure}[t]
    \centering
    \includegraphics{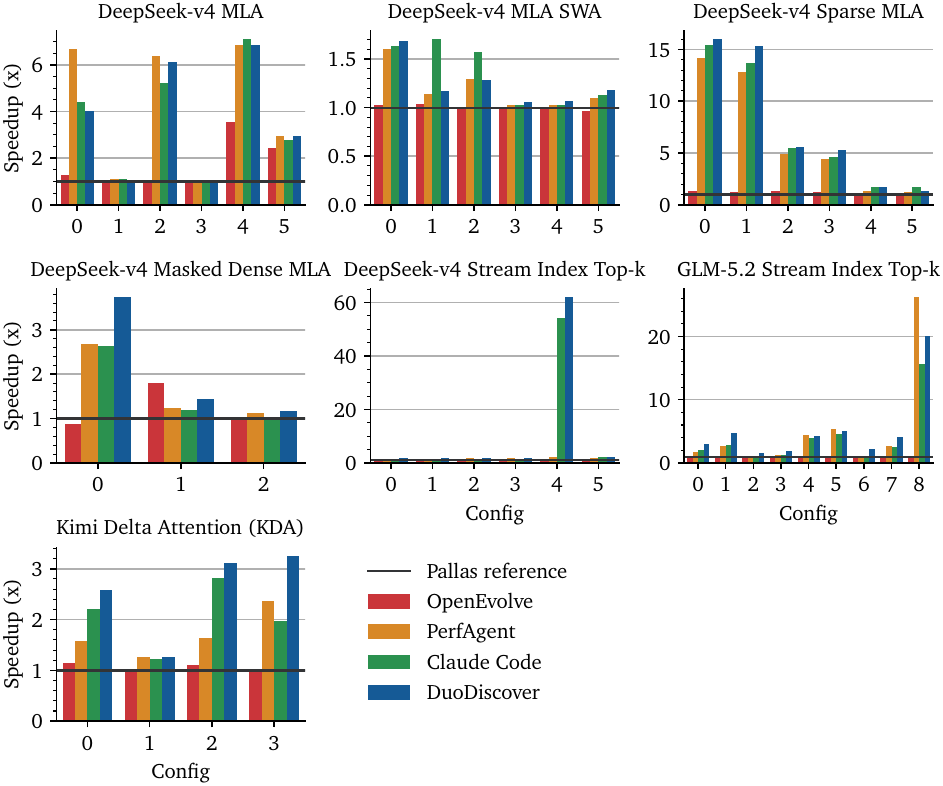}
    \caption{Speedup over the optimized production Pallas reference (horizontal line at 1.00$\times$) on every evaluated configuration of each \tpuinf kernel, for \sys, Claude Code, \perfagent, and \openevolve.}
    \label{fig:tpu_inference_bench}
\end{figure}

The \tpuinf results cover 7 production TPU kernels provided by the vLLM TPU-inference project~\citep{vllm_tpu_inference}, including DeepSeek-V4 multi-head latent attention (MLA) variants, DeepSeek-V4 and GLM-5.2 stream-index top-$k$, and Kimi Delta Attention (KDA). The speedup obtained on each configuration of every kernel is plotted in Figure~\ref{fig:tpu_inference_bench}. The configurations include representative prefill and decode inference settings. These kernels were chosen after correspondence with the vLLM TPU-inference team because they cover a variety of important kernels from recent open frontier models and have mature testing and benchmarking scripts available. Note that, at the time of writing, the KDA kernel is not part of the public TPU-inference GitHub repository yet. The kernel was provided to us through correspondence with the vLLM TPU-inference team. The reference implementation for each kernel is the optimized production Pallas kernel, so a speedup of 1.00$\times$ means no improvement over already-tuned code; unlike \jax, the denominator is not XLA, and the two suites should not be compared. Every method ran with a 10-hour budget and four agents (for \sys, three explorers and one control). Each kernel is scored by the geometric mean over its configurations, and the reported score is the geometric mean over the 7 kernels, the same convention as \jax: \grounded{tpuinf.duodiscover}{2.68}$\times$ for \sys, \grounded{tpuinf.claudecode}{2.38}$\times$ for Claude Code, \grounded{tpuinf.perfagent}{2.14}$\times$ for \perfagent, and \grounded{tpuinf.openevolve}{1.14}$\times$ for \openevolve (Table~\ref{tab:main}). \sys is first on 5 of the 7 kernels. \perfagent is ahead on DeepSeek-V4 MLA and Claude Code on DeepSeek-V4 MLA SWA.

\section{Case study details}\label{app:deployment_details}

\begin{figure}[t]
\centering
\includegraphics{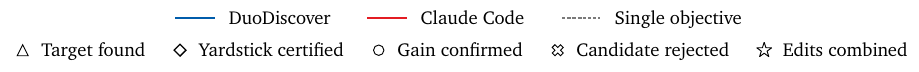}\\[3pt]
\begin{subfigure}{0.48\linewidth}
\centering
\includegraphics[width=\linewidth]{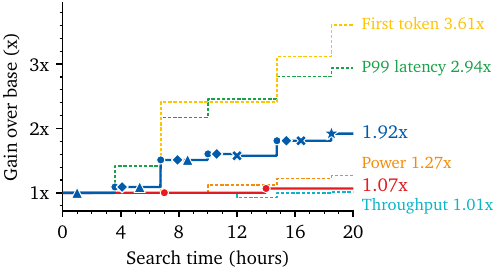}
\caption{vLLM on an 8$\times$B200 node, 20 h budget each. \sys 7 true evaluations (9.9 h of evaluation time), Claude Code 12 (16.8 h).}
\label{fig:case-vllm-detail}
\end{subfigure}\hfill
\begin{subfigure}{0.48\linewidth}
\centering
\includegraphics[width=\linewidth]{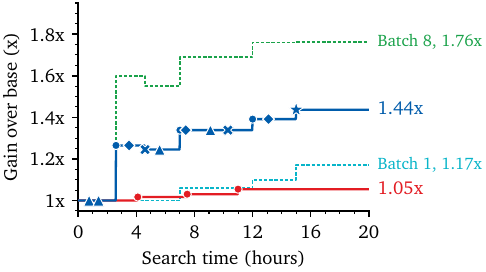}
\caption{FreeToken on a 48 GB L40S, 20 h budget each. \sys 6 true evaluations (6.9 h of evaluation time), Claude Code 14 (16.3 h).}
\label{fig:case-freetoken-detail}
\end{subfigure}
\caption{Gain over the base code as the search runs. Thick solid lines are the geometric mean over the objectives at each true evaluation, \sys in blue and Claude Code in red. The main text shows the same curves without event markers (Figure~\ref{fig:casestudies}). Thin dashed lines are the individual objectives behind that mean. Markers on the thick line are the search events in the legend. For Claude Code only confirmed gains are marked.}
\label{fig:casestudies-detail}
\end{figure}

This appendix expands \S\ref{sec:deployment}. The runs are those of \S\ref{sec:deployment}, the budgets are in Table~\ref{tab:benchmarks}, and the baseline is four Claude Code agents on Opus 5 as in Table~\ref{tab:baselines}. The data-prefetcher study is the exception, comparing against ArchAgent v2 and the best human design (Table~\ref{tab:prefetcher}). The fidelity test's floor is two evaluated states, the original code and the first candidate, instead of the six used on the benchmarks, since a target sees only a few true evaluations of 10 to 85 minutes each, and up to 28 hours for the prefetcher.

\subsection{vLLM}\label{app:vllm}
\begin{itemize}
\item \textbf{Setting.} vLLM serves GLM-5.2-NVFP4 on one 8$\times$B200 node, tensor parallel over the eight GPUs with expert parallelism and an fp8 key-value (KV) cache. The true evaluation scores p99 decode latency, first-token latency, power and throughput and takes their geometric mean. The baseline is unmodified vLLM tuned for the workload, with priority scheduling on, a long-prefill token threshold set, and at most 64 sequences in flight. A true evaluation sweeps a workload of 8192 input and 1024 output tokens at rising arrival rates, samples power once a second, requires the test suite to pass, and checks that greedy output equals the unmodified build's. It takes about 85 minutes.
\item \textbf{Search.} Figure~\ref{fig:case-vllm-detail} shows the 20 hours. All four agents started on throughput until one profiled the server and found decode bound by memory latency. Streaming multiprocessor (SM) issue was at 7.8\% and DRAM reads at 21\% of peak. The orchestrator dropped the throughput targets, terminated one agent, and briefed the other three to explore. They found targets in p99 decode latency, GPU power, and first-token latency for a light client while a heavy client runs, all at fixed throughput. A decode cost model scored the prefill cap in seconds, and a 13-minute load sweep scored the per-client fair scheduler and the clock control. Each yardstick passed the fidelity test after its target's first true evaluation (the diamonds). The test invalidated the load sweep when higher load reversed its ranking, and at hour 15.4 validated the sweep rebuilt on two arrival rates. At hour 7 the orchestrator spawned a fourth agent onto the first-token target and handed it the p99 agent's validated prefill cap, re-tested on arrival. The fourth agent's queue change, confirmed at hour 14.75, lifted the first-token gain from \diffspeedup{2.4} to \diffspeedup{3.1}. Of about 90 candidates, 7 went to true evaluation, 9.9 hours of evaluation time in the 20-hour run. Two candidates were rejected (the crosses). At hour 12 a prefill cap set below the decode floor (the step time of decode alone) cost 7\% of throughput. At hour 16.4 a shortest-job-first schedule turned out to be one the unmodified server already provides. Prefix co-scheduling and preemptive fairness failed on their yardsticks and never cost a true evaluation. The seventh true evaluation, at hour 18.5, confirmed the combined edits (the star). Claude Code kept every agent on throughput for 10 hours, then moved to first-token latency. It confirmed two gains, throughput 1.01$\times$ at hour 7 and first-token latency 1.29$\times$ at hour 14, with p99 decode latency and power unchanged, and ended at \diffspeedup{1.07} after 12 true evaluations, 16.8 hours of evaluation time, against \sys's \diffspeedup{1.92}.
\item \textbf{Solution.} The solution is about 500 lines in four pieces. None of it touches a kernel, and greedy output equaled the unmodified build's byte for byte on all six check prompts. \texttt{scheduler.py} caps the prefill tokens per step from the cost model, so a decode step cannot stall behind a long prefill. The cap is recomputed each step from the current decode batch. The unmodified server's per-step limit, max-num-batched-tokens, is static and cannot do that. \texttt{dvfs\_governor.py} lowers the GPU clock while decode latency has slack, which cuts power. A per-client fair scheduler, a virtual token counter, gives each client a weighted share of served tokens, so the light client's first token no longer waits behind the heavy client's backlog. The unmodified scheduler's priority is static and per request and gives no such share. One flag, the decode tail latency to hold, sets the prefill cap and the clock together. At hour 20 p99 decode latency is \diffspeedup{2.94} lower, first-token latency \diffspeedup{3.61} lower, and power 21\% lower, a geometric mean of \diffspeedup{1.92} over the four objectives with throughput at 1.01$\times$. Table~\ref{tab:vllm_results} uses fixed arrival rates rather than the true evaluation's rising-rate sweep, so its numbers exceed the hour-20 ones. The throughput row is the full solution at hour 20. The p99 and power rows use 1.0 and 1.5 requests per second. The first-token row compares the fair scheduler with first come first served. The tokens-per-joule row holds the clock at 1200 MHz against 1965 MHz on a short-prompt load, where the clock alone cuts power 24.9\% and throughput 1.3\%. The concurrent-prefill row compares the per-step cap with the unmodified server's long-prefill threshold at 256 or 2048 tokens. The clock control lowers the tail as well as power. A lower clock keeps more decodes in flight per step, so prefill is spread thinner and the collision spikes that set the tail shrink. The p99 then sits just above the decode floor set by the clock, so the cap and the clock control do not stack on tail latency. Together they give 23.1 ms p99 on the served workload, the same as the clock control alone, and the cap alone at full clock gives 16.5 ms. The pair gives up 6.6 ms of tail latency for the table's 34\% power cut at equal throughput. Against the unmodified server's default chunked prefill at 16,384 batched tokens per step, the cap is a worst-case tail guard. It cuts median decode latency 10 to 15\% at some decode counts and leaves it unchanged at others. Its gain is the p99 row of Table~\ref{tab:vllm_results}.
\end{itemize}

\begin{table}[t]
\centering\footnotesize
\setlength{\tabcolsep}{3pt}
\caption{Each mechanism of the vLLM solution measured after the run on 8$\times$B200 serving GLM-5.2-NVFP4, under the fixed-rate conditions given in the text. The first-token row is against first-come-first-served scheduling, the throughput row is the full solution.}
\label{tab:vllm_results}
\begin{tabularx}{\linewidth}{>{\raggedright\arraybackslash}X >{\raggedright\arraybackslash}p{0.2\linewidth} >{\raggedright\arraybackslash}p{0.17\linewidth} >{\raggedright\arraybackslash}p{0.12\linewidth}}
\toprule\addlinespace[1pt]
Measure & Tuned unmodified vLLM & \mbox{\systab} & Change \\
\midrule
p99 decode latency (inter-token latency), real workload & 141 to 337 ms & 22 to 26 ms & \xspeedup{6}{13} lower \\
GPU power at equal throughput & 4594 to 4968 W & 3038 to 3237 W & 34\% lower \\
Tokens per joule, fixed low clock & baseline & 1.314$\times$ baseline & 31.4\% higher \\
Light-client first-token latency under a heavy client (vs.\ first come first served) & 9654 ms & 1114 ms & \diffspeedup{8.7} lower \\
p99 decode latency under 8 concurrent long prefills & 221.6 to 507.7 ms (baseline cap) & 24.3 ms & \xspeedup{9}{21} lower \\
Peak throughput & baseline & unchanged & within 1\% \\
\bottomrule
\end{tabularx}
\end{table}

\subsection{FreeToken}\label{app:freetoken}
\begin{itemize}
\item \textbf{Setting.} FreeToken runs GLM-5.2-NVFP4, a 753B mixture-of-experts model with 40B active parameters, on one 48 GB L40S. The model's 433 GB of weights live in host memory, with a cache of hot experts on the GPU. Each decode step fetches a fixed fraction of the experts it needs over PCIe and computes the rest on the CPU. Decode is bound by those fetches and by the CPU expert path. A true evaluation builds the engine and decodes 256 tokens at batch 1 and batch 8, three times each. The test suite must pass and greedy output must equal the unmodified build's on the evaluation's prompts. It scores decode throughput at batch 1 and batch 8 and takes their geometric mean. A true evaluation takes about 70 minutes.
\item \textbf{Search.} Figure~\ref{fig:case-freetoken-detail} shows the 20 hours. Each agent explored alone, chose a target, and built its own yardstick, a two-minute microbenchmark or a byte-count model. Two converged on the CPU expert loop, so the orchestrator terminated one and spawned a new agent, which found the shared-expert schedule. A third agent's target was the fp8 linear layer, scored against the floor set by the bytes it must move. A fourth's was the expert cache policy, scored by the bytes its routing sends over PCIe. The CPU-loop, shared-expert, and fp8 yardsticks passed the fidelity test at hours 3.5, 7.4, and 13.1 (the diamonds), each after a true evaluation on its target. At hour 4.6 a true evaluation rejected the \diffspeedup{1.25} that the cache-policy yardstick had credited to a cut of a few percent in expert fetches. That yardstick was dropped (the first cross). Of about 80 candidates, 6 went to true evaluation, 6.9 hours of evaluation time in the 20-hour run. Gains were confirmed at hours 2.6, 7, and 12 (the circles). A tuned fetch fraction at hour 10.3 gave no gain over the confirmed best (the second cross), and the combined edits were confirmed at hour 15 (the star). Claude Code rewrote the batch-8 kernels and ended at \diffspeedup{1.05} after 14 true evaluations, 16.3 hours of evaluation time, against \sys's \diffspeedup{1.44}.
\item \textbf{Solution.} \texttt{cpu\_moe\_ext.cpp} visits each expert's weights once while they are still in cache. The CPU expert loop is the largest gain. The shared expert runs inside the window where the GPU would otherwise wait for the CPU. \texttt{fp8\_pertensor\_linear.py} had a dispatch rule, written when decode meant one token, that sent every batch above one to the general tiled GEMM. At batch 8 that grid is 48 blocks of four warps on a 142-SM GPU, so most SMs sit idle. One added branch routes those batches to a new split-K kernel that fills the GPU as the batch-1 path does. Per batch-8 decode step the layer's calls fall from 38.2 to 24.1 ms. That 14.1 ms is most of the 19 ms cut from the 469 ms step. The 24.1 ms is close to the 22.9 ms the batch-1 path takes to move the same bytes. With this kernel on top of the other edits, the true evaluation's batch-8 ratio went from \diffspeedup{1.69} to \diffspeedup{1.76}, three true evaluations within 0.7\% of each other. The rule returned correct values on every branch, so no test caught the misroute, and the byte-floor yardstick showed how far above the floor the layer ran. An early skip for the expert routes that the mask already zeroes removes about 37\% of the GPU expert-GEMV memory reads. A worker loop that spins briefly before it parks removes a wakeup stall per layer. The masked-route skip and the shared-expert overlap are bit-identical by construction. The others passed the greedy-output gate. The final engine also carries edits confirmed as part of the combined edits at hour 15. Hot experts stay GPU-resident under a least-frequently-used policy with recency decay. Fetches are ordered by how many tokens need each expert, host-to-device copies overlap CPU compute, and the PCIe fetch fraction is set per model and GPU. At batch 1 each expert serves one token, so visiting its weights once saves nothing, and the batch-1 gain comes from the other edits. Table~\ref{tab:freetoken_results} gives the confirmed gain at each batch size, a geometric mean of \diffspeedup{1.44}.
\end{itemize}

\begin{table}[t]
\centering\footnotesize
\setlength{\tabcolsep}{6pt}
\caption{FreeToken on one L40S, decode tokens per second against the unmodified build, measured by the true evaluation, three runs each.}
\label{tab:freetoken_results}
\begin{tabular}{l c r r c}
\toprule
Model & Batch & Unmodified & \systab & Speedup \\
\midrule
GLM-5.2-NVFP4 (753B, 40B active) & 1 & 6.75 & 7.9 & \diffspeedup{1.17} \\
GLM-5.2-NVFP4 (753B, 40B active) & 8 & 10.10 & 17.8 & \diffspeedup{1.76} \\
\bottomrule
\end{tabular}
\end{table}

\subsection{SQLite}\label{app:sqlite}
\begin{figure}[t]
\centering
\includegraphics{figs/fig_case_legend.pdf}\\[3pt]
\begin{subfigure}{0.48\linewidth}
\centering
\includegraphics{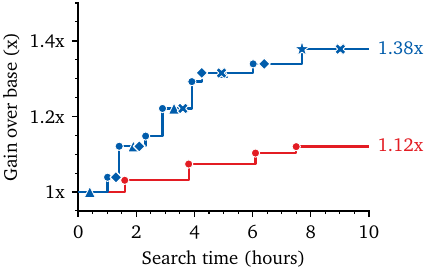}
\caption{SQLite, wall time, 10 h budget each. \sys 17 true evaluations (2.8 h of evaluation time), Claude Code 45 (7.5 h).}
\label{fig:case-sqlite}
\end{subfigure}\hfill
\begin{subfigure}{0.48\linewidth}
\centering
\includegraphics{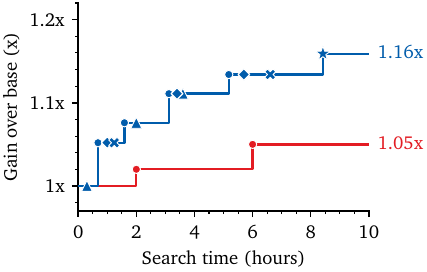}
\caption{zstd, instruction count, 10 h budget each. \sys 7 true evaluations (2.4 h of evaluation time), Claude Code 21 (7.3 h).}
\label{fig:case-zstd}
\end{subfigure}
\caption{Gain over the base code as the search runs, for SQLite and zstd, each a single objective, the zstd axis an instruction-count ratio. Lines and markers as in Figure~\ref{fig:casestudies-detail}.}
\label{fig:case-progbench}
\end{figure}
\begin{itemize}
\item \textbf{Setting.} The task is to make SQLite's own \texttt{speedtest1} workload faster, editing only \texttt{src/}. A true evaluation rebuilds the engine, runs the 13,249-test suite twice, times \texttt{speedtest1} at size 750 three times, and checks a result hash on seeds and query shapes the agent never sees. It takes about 10 minutes.
\item \textbf{Search.} The agents found four targets in turn, the floating-point paths of the bytecode engine, the sorter, the recursive query path, and the decimal conversion (the triangles in Figure~\ref{fig:case-sqlite}). Each target used the same yardstick, the \texttt{cte} set of \texttt{speedtest1} at sizes 100 and 200, a recursive query over about 43 million rows that runs in seconds, tested separately per target. The four fidelity tests passed at hours 1.3, 2.1, 4.3, and 6.4 (the diamonds), each after a true evaluation on its target. Their scores stayed within 2 to 4\% of the true evaluations. Of about 200 candidates, 17 went to true evaluation, 2.8 hours of evaluation time in the 10-hour run. Gains were confirmed at hours 1.0, 1.4, 2.3, 2.9, 3.9, 4.3, and 6.0 (the circles). The combined edits were confirmed at hour 7.7 (the star). Two candidates the yardstick ranked above the best confirmed candidate scored below it at hours 3.6 and 4.9, and one broke a test at hour 9.0 (the crosses). The other six scored below the best confirmed candidate and are not marked. Two agents converged on one target, and the kept edit was transferred from an earlier confirmed one. Claude Code ran 45 true evaluations, 7.5 hours of evaluation time, and ended at \diffspeedup{1.12}, against \sys's \diffspeedup{1.38}.
\item \textbf{Solution.} The bytecode engine had fast paths for integer arithmetic, comparison and sorting but not for floats. \texttt{vdbe.c}, \texttt{vdbeaux.c}, and \texttt{vdbesort.c} add the same paths for floats, and the sorter merges 512 runs per pass instead of 16. \texttt{expr.c} and \texttt{select.c} compute a repeated subexpression once and, for a recursive query with no ORDER BY, pop the newest row from the queue first, a depth-first traversal that SQL leaves unspecified. \texttt{util.c} converts a decimal with a single correctly rounded multiply or divide when the mantissa is at most $2^{53}$ and the power of ten is within $\pm$22. \texttt{pcache1.c} computes its page-cache hash index with a bit mask instead of the original modulo. Two further edits are build settings inside the editable \texttt{src/}: \texttt{sqliteInt.h} adds an O3 optimize pragma, and \texttt{global.c} sets the default mutexes off, which the single-threaded workload never contends. Both passed the 13,249-test suite twice and the true evaluation. The confirmed gain is \diffspeedup{1.38}, with the result hash equal to the unmodified build's.
\end{itemize}

\subsection{zstd}\label{app:zstd}
\begin{itemize}
\item \textbf{Setting.} The task is to make zstd 1.6.0 execute fewer instructions on a fixed workload while producing the same bytes, editing only \texttt{lib/}. The task scores instruction count, so we report an instruction ratio, not a speedup. A true evaluation runs the 2,038-test suite over 14 branches and counts instructions under callgrind, Valgrind's instruction-count profiler, across 16 cells, four files from the Silesia compression corpus at levels 3, 9, 15, and 19. The score is the geometric mean over cells of the unmodified build's count over the candidate's. Any difference from the unmodified build's bytes scores 0. A true evaluation takes 15 to 25 minutes.
\item \textbf{Search.} The agents found two targets, the optimal parser in \texttt{compress/zstd\_opt.c} that level 19 uses and the row hash match finder in \texttt{compress/zstd\_lazy.c} that level 9 uses. A third came later, the 32-byte match compare (the triangles in Figure~\ref{fig:case-zstd}). The yardstick for each target took seconds, counting instructions under callgrind on one small file at one level. The parser's yardstick passed the fidelity test at hour 1.0, the match finder's at hour 3.4, and the compare's at hour 5.7 (the diamonds), each after a true evaluation on its target. The parser yardstick's score was within 0.04\% of the first true evaluation. Of about 90 candidates, 7 went to true evaluation, 2.4 hours of evaluation time in the 10-hour run. Gains were confirmed at hours 0.7, 1.6, 3.1, and 5.2 (the circles). The combined edits were confirmed at hour 8.4 (the star). Two candidates the yardstick ranked above the best confirmed candidate scored below it at hours 1.2 and 6.6 (the crosses). The orchestrator transferred two of the final edits from a second agent. Claude Code ran 21 true evaluations, 7.3 hours of evaluation time, and ended at \diffspeedup{1.05}, against \sys's \diffspeedup{1.16}.
\item \textbf{Solution.} \texttt{compress/zstd\_opt.c} builds the length and offset price tables once per forward pass instead of repricing every match. \texttt{compress/zstd\_lazy.c} compares each match candidate as it is found instead of collecting and revisiting them. \texttt{compress/zstd\_compress\_internal.h} compares 32 bytes at a time with AVX2, \texttt{common/compiler.h} removes the software prefetches, and 13 compressor files gain an AVX2 \texttt{\#pragma GCC target} line. The prefetch removal and single-pass finder cut the instruction count, and the AVX2 pragma compiles the hot files for AVX2. The confirmed gain is \diffspeedup{1.16} with identical bytes.
\end{itemize}

\subsection{Data-prefetcher design}\label{app:prefetcher}
\begin{itemize}
\item \textbf{Setting.} The task is the 4th Data Prefetching Championship (DPC4)~\citep{bera2026dpc4} on the ChampSim simulator~\citep{gober2022championship}. 
The baseline setup and starting point is a ChampSim single- or multi-core system with a multi-level prefetcher; Berti~\citep{navarro2022berti} at the L1D, Pythia~\citep{bera2021pythia} at the L2, and nothing at the LLC, with a storage limit of 32, 128, and 256\,KB per cache level. 
Both Berti and Pythia were the best prior published prefetchers. 
A design is scored by its geometric-mean speedup in instructions per cycle (IPC) over the reference core in three configurations: 
FullBW runs one core at full bandwidth, LimitBW runs one core with a throttled memory bus,
and MC runs four cores sharing memory. The overall score is the geometric mean of the three. 
FullBW and LimitBW run 610 traces split among SPEC CPU2017, Google~\citeyearpar{Google_Workload_Traces_Version_2}, AI/ML, and graph analytics workload suites. MC runs 484 mixes of four traces each from the same suites used for single-core measurements: 384 take all four traces from one suite, and 100 take one trace from each suite.
A true evaluation runs about 1{,}700 simulations (610 traces each on FullBW and LimitBW, and 484 mixes on MC) and totals about 2{,}600 CPU-hours (5 to 9 hours of wall-clock for FullBW and LimitBW, and 25 to 28 hours for MC, running at most 120 simulations at a time per configuration),
so the agents hill-climb yardsticks built from short subsets of the traces, and only the best designs reach a true evaluation.
\item \textbf{Search.}
The search ran for 16.7 days as a sequence of 12-hour runs with three agents each, every run seeded from the previous best. The best design of each run, by its score on the 131 released traces, seeded the next run; the official evaluations ran asynchronously in parallel alongside subsequent runs and did not gate them.
The search was conducted by the agents autonomously, except for a few targeted interventions.
The agents started on the L1D prefetcher alone, scored on FullBW. Those designs improved FullBW but lost most of their gain once the memory bus was throttled, so on day 4 we changed the search score to the geometric mean of FullBW and LimitBW, which penalizes wasted prefetches. On day 6 we let the agents edit the L2 prefetcher as well. The scoring harness also accepted LLC changes from then on, and on day 8 an agent, unprompted, wrote the first LLC prefetcher, a temporal correlator. 
MC stayed flat until we added 15 four-core mixes built from released traces to the search score for the final run. 
The agents' yardstick was single-core only, so each agent wrote its own four-core yardstick over those mixes.
One agent keyed the LLC correlator by core, another found that the L2's page-table engine was hurting four-core runs and switched it off there, and the orchestrator had a third agent merge the two.
Together these changes raised MC from 1.0274$\times$ to 1.0393$\times$, with FullBW and LimitBW unchanged.
Of about 8{,}000 candidates scored on the 131 released traces, 38 reached a true evaluation on FullBW, 27 on LimitBW and 18 on MC.
\item \textbf{Solution.} The final design is about 5{,}900 lines of C++. The L1D is a rewritten Berti with deeper cross-page lookahead and latency-triggered throttling. The L2 extends Pythia with an accuracy gate, two footprint engines, and a page-table engine that is off on four cores. The last-level cache has a temporal correlator, keyed by core on four cores. It reaches a 1.0631$\times$ overall speedup against ArchAgent v2's 1.0382$\times$ and the best human design's 1.0352$\times$ (Table~\ref{tab:prefetcher}).
\end{itemize}

\section{Ablation and scaling details}\label{app:ablations}
This appendix expands \S\ref{sec:ablations} with a single task-wide evaluator in place of per-target yardsticks, the fleet-size sweep, and the fleet-scaling curves over time.

\subsection{Per-target yardsticks against a single task evaluator}\label{app:one_evaluator}
\begin{figure}[t]
\centering
\begin{minipage}[c]{0.5\linewidth}
\centering\footnotesize
\setlength{\tabcolsep}{4pt}
\resizebox{\linewidth}{!}{\begin{tabular}{l l c c c}
\toprule
& & Score & Validated & Cost \\
\midrule
\sweeff & \systab & 14.93$\times$ & 58\% & 4\% \\
& one evaluator & 12.98$\times$ & 21\% & 29\% \\
\solexec & \systab & 80.80\% & 73\% & 2\% \\
& one evaluator & 71.36\% & 33\% & 18\% \\
\bottomrule
\end{tabular}}
\captionof{table}{A yardstick per target against one evaluator per task. Validated is the share of targets (\sys rows) or tasks (one-evaluator rows) whose yardstick passes the fidelity test, Cost the run time over one true evaluation.}
\label{tab:one_evaluator}
\end{minipage}\hfill
\begin{minipage}[c]{0.47\linewidth}
\centering
\includegraphics{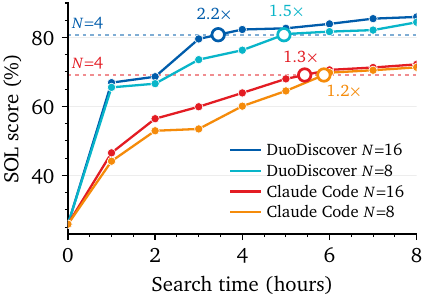}
\captionof{figure}{8- and 16-agent fleets over 8 hours on the 50 \solexec kernels. Dotted lines mark each method's four-agent result, rings where each fleet crosses it, labeled with the wall-clock speedup.}
\label{fig:fleet_time}
\end{minipage}
\end{figure}
Instead of a yardstick per target, an agent could build one fast evaluator for the whole task and hill-climb on it. We test this design on the 20 repository tasks and 50 kernels of Section~\ref{sec:ablations} with everything else unchanged (Table~\ref{tab:one_evaluator}). A task-wide evaluator cannot be both fast and correlated with the true evaluation.

On the repositories it passes the fidelity test on 21\% of tasks and, where validated, costs 29\% of a true evaluation. The score drops from \diffspeedup{14.93} to \diffspeedup{12.98} on repositories and by 9.44 in SOL score, from 80.80\% to 71.36\%, on kernels. Amdahl's law~\citep{amdahl1967} explains the breakdown. A yardstick only has to rank candidates within one region of the code. A task-wide evaluator must rank edits from every region, and to run fast it must trim its workload. An edit's gain grows with the share of the measured workload's time its region takes, so a trimmed workload divides time across regions differently from the real one, and the evaluator misranks edits from different regions relative to the true evaluation.

\subsection{Fleet scaling}\label{sec:scaling}\label{sec:generality}

\begin{wrapfigure}{r}{0.42\linewidth}
\centering
\vspace{-12pt}
\includegraphics[width=\linewidth]{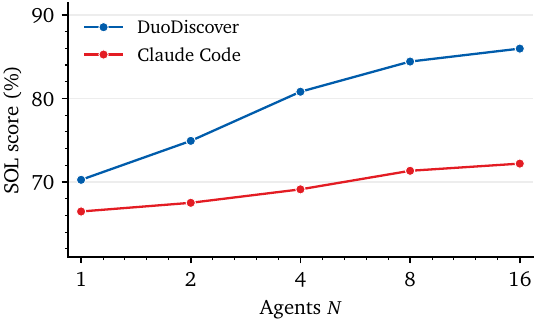}
\caption{SOL score against agent count $N$, 8-hour runs on 50 \solexec kernels.}
\label{fig:fleet_scaling}
\vspace{-10pt}
\end{wrapfigure}
To test fleet scaling, we evaluate \sys and \claudecode across 50 \solexec kernels for fleet sizes $N \in \{1, 2, 4, 8, 16\}$ under equal wall-clock time (Figure~\ref{fig:fleet_scaling}). At $N=1$, \sys attains 70.26\% SOL versus 66.47\% for \claudecode due to higher candidate throughput on validated yardsticks. While the baseline rises slowly to 72.20\% at $N=16$ as isolated agents duplicate each other's search, \sys climbs to 84.41\% at $N=8$ and 85.96\% at $N=16$ as orchestration transfers discoveries across agents, so the gap widens at every fleet size, from 3.8 to 13.8 SOL. At $N=16$, \sys attains its 4-agent benchmark score 2.2$\times$ faster (3.4\,h vs. 7.6\,h), compared to 1.3$\times$ for \claudecode (Appendix~\ref{app:fleet_time}).

\subsection{Fleet scaling over time}\label{app:fleet_time}
Figure~\ref{fig:fleet_time} tracks the 8- and 16-agent fleets over the 8-hour budget. Each curve starts at the reference implementation's score of about 26\% and climbs to that method's final 8-hour score. On the 50 \solexec kernels, a larger fleet buys \sys far more time than it buys Claude Code. With 16 and 8 agents, \sys attains its four-agent result of 80.80\% at 3.4 and 5.0 hours. The four-agent run itself gets there at 7.6 hours, so the larger fleets are 2.2$\times$ and 1.5$\times$ faster. Claude Code needs 5.4 hours with 16 agents and 5.9 hours with 8 for its 69.12\%, against 6.8 hours with four, so its larger fleets are 1.3$\times$ and 1.2$\times$ faster. The four-agent times are the hours at which the four-agent curves of Figure~\ref{fig:main_tradeoff}a settle at their final scores, and the speedups are computed before rounding. The difference comes from the orchestrator transferring discoveries across agents, where isolated agents duplicate each other's search.

\section{Yardstick and search analysis}\label{app:analysis}
This appendix expands Section~\ref{sec:ablations}. It gives yardstick speed and fidelity per target, how a run finds targets and hill-climbs, the token cost of a run, four examples of discovered targets and yardsticks, and the stress tests and run-to-run spread.

\subsection{Yardstick quality}\label{app:yardstick_quality}
Table~\ref{tab:yardstick_quality} (Section~\ref{sec:ablations}) gives the speed and rank fidelity of the yardsticks agents build on the 20 \sweeff tasks and 50 \solexec kernels. Fidelity is Kendall's rank correlation $\tau$ between a yardstick's scores and the true evaluation's over the candidates that yardstick sent to a true evaluation, reported as the median over validated yardsticks. The fidelity test validates a yardstick at $\tau \ge 0.5$ over at least six resolved states (\S\ref{sec:fidelity}). The test reuses states that any agent sharing the yardstick has already truly evaluated, so one true evaluation serves several tests, which is how a kernel run's 54 true evaluations cover its 11 validated yardsticks.

\begin{figure}[t]
\centering
\includegraphics{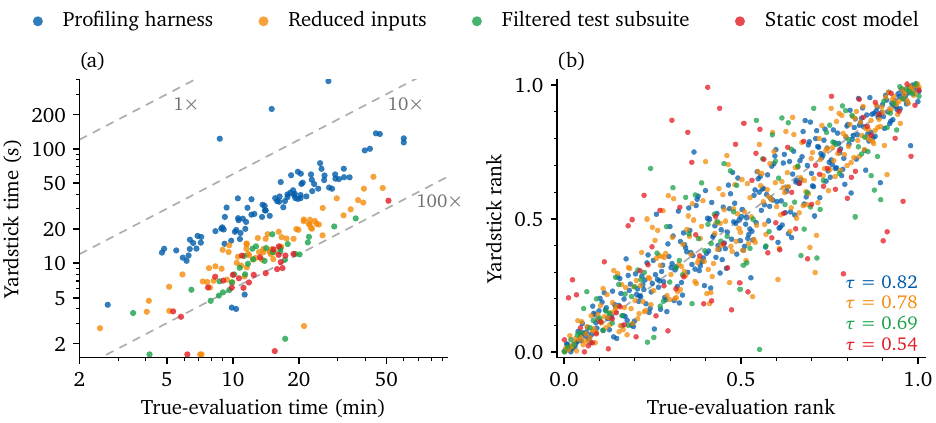}
\caption{Yardstick quality by kind on 15 of the 50 \solexec kernels. (a) Yardstick time against true-evaluation time, one point per target, dashed lines at 1, 10 and 100$\times$ faster. (b) Rank correlation between yardstick and true evaluation, one point per truly evaluated candidate, its rank by the yardstick against its rank by the true evaluation, pooled over the candidates of each kind, with Kendall $\tau$ per kind.}
\label{fig:yardstick_quality}
\end{figure}

Two selection effects pull $\tau$ in opposite directions. True evaluations go only to candidates the yardstick favors, which narrows the score range and lowers $\tau$, and only validated yardsticks are counted, which raises it. Over the kernel yardsticks that were never validated, the median is 0.03. A second view needs neither selection, the share of truly evaluated candidates that turn out to be gains. On kernels, 37.0\% (20 of 54) are gains found on a yardstick, plus 1 combined edit, against 21.1\% (16 of 76) for Claude Code. On repositories, 17.5\% (10.3 of 59) are, plus about 2 combined edits, against 8.6\% (6.2 of 72). Per confirmed gain, \sys spends 2.6 true evaluations on kernels and 4.8 on repositories, against 4.8 and 11.6 for Claude Code. Figure~\ref{fig:yardstick_quality} plots per-target speed and, for every truly evaluated candidate, its yardstick rank against its true-evaluation rank on 15 of the 50 kernels. Its $\tau$ is pooled over all candidates of a kind, so it differs from the per-yardstick median in the table and from the fidelity test, which computes $\tau$ per target over noise-resolved pairs (\S\ref{sec:fidelity}).

\subsection{Search behavior}\label{app:search_analysis}
The run logs of the 50 \solexec kernels and 20 \sweeff tasks show how agents discover targets, build yardsticks and hill-climb. All counts are means over the 50 kernels or the 20 tasks unless a median is named.

\textbf{Target and yardstick co-discovery.} Agents pick their own targets and build a yardstick for each, with no manual rule. A run pursues 15 targets per kernel, 11 of which get a validated yardstick, against 3 to 7 targets for Claude Code (Table~\ref{tab:search_summary}). On repository tasks it pursues 9.2 targets per task, 5.3 validated, against Claude Code's 2 to 4. Two of the 20 tasks end with no validated yardstick. The yardsticks fall into four kinds (Table~\ref{tab:yardstick_kinds}), the same four on both benchmarks with different shares: reduced inputs, a profiling harness, a filtered test subsuite, and a static cost model. The filtered test subsuite times the tests on the changed code, so it scores speed like the others, and it also has to keep those tests passing. Static cost models are the fastest kind and the least often validated, 32\% on kernels and 18\% on repositories, and one found the fp8 dispatch misroute in FreeToken that no test could catch (Appendix~\ref{app:freetoken}).

\begin{table}[t]
\centering\footnotesize
\setlength{\tabcolsep}{4pt}
\caption{The four kinds of yardstick agents build on the 50 \solexec kernels and the 20 \sweeff tasks. Weighted by share, the kernel columns give the 53$\times$ and 11 of 15 of Table~\ref{tab:yardstick_quality}, the repository columns the 22$\times$ and 5.3 of 9.2.}
\label{tab:yardstick_kinds}
\begin{tabularx}{\linewidth}{l >{\raggedright\arraybackslash}X c c c c c c}
\toprule
& & \multicolumn{3}{c}{\solexec kernels} & \multicolumn{3}{c}{\sweeff tasks} \\
\cmidrule(lr){3-5} \cmidrule(lr){6-8}
Yardstick kind & Scores a candidate on & Share & Faster by & Validated & Share & Faster by & Validated \\
\midrule
Reduced inputs         & smaller or low-precision inputs & 33\% & 63$\times$ & 75\% & 42\% & 20$\times$ & 67\% \\
Profiling harness      & an isolated hot loop            & 42\% & 28$\times$ & 81\% & 29\% & 11$\times$ & 58\% \\
Filtered test subsuite & tests on the changed code       & 14\% & 78$\times$ & 69\% & 24\% & 31$\times$ & 51\% \\
Static cost model      & instruction or memory counts    & 11\% & 89$\times$ & 32\% & 5\% & 52$\times$ & 18\% \\
\bottomrule
\end{tabularx}
\end{table}

\begin{table}[t]
\centering\footnotesize
\setlength{\tabcolsep}{5pt}
\caption{Search per task, means over 20 \sweeff tasks and 50 \solexec kernels, times are medians. Ratio is \sys over Claude Code, arrow up for more or faster, down for fewer true evaluations. Confirmed gains include the combined edits. Gains per true evaluation counts only edits found on a yardstick.}
\label{tab:search_summary}
\begin{tabularx}{\linewidth}{>{\raggedright\arraybackslash}X rr>{\color{gray}}r rr>{\color{gray}}r}
\toprule
& \multicolumn{3}{c}{\sweeff (20 tasks)} & \multicolumn{3}{c}{\solexec (50 kernels)} \\
\cmidrule(lr){2-4}\cmidrule(lr){5-7}
& Claude Code & \systab & ratio & Claude Code & \systab & ratio \\
\midrule
Targets pursued           & 2 to 4 & \textbf{9.2}  & -- & 3 to 7 & \textbf{15} & -- \\
Yardsticks validated        & --     & \textbf{5.3}  & -- & --     & \textbf{11} & -- \\
Time per candidate check & 15 min & \textbf{40 s} & 22$\times\,\uparrow$ & 14 min & \textbf{16 s} & 53$\times\,\uparrow$ \\
\specialrule{0.2pt}{2pt}{2pt}
Candidates scored         & 72     & \textbf{358}  & 5.0$\times\,\uparrow$ & 76 & \textbf{285} & 3.8$\times\,\uparrow$ \\
True evaluations          & 72     & \textbf{59}   & 0.8$\times\,\downarrow$ & 76 & \textbf{54}  & 0.7$\times\,\downarrow$ \\
\specialrule{0.2pt}{2pt}{2pt}
Confirmed gains           & 6.2    & \textbf{12.3} & 2.0$\times\,\uparrow$ & 16 & \textbf{21} & 1.3$\times\,\uparrow$ \\
Gains per true evaluation & 8.6\%  & \textbf{17.5\%} & 2.0$\times\,\uparrow$ & 21.1\% & \textbf{37.0\%} & 1.8$\times\,\uparrow$ \\
Score                     & 12.85$\times$ & \textbf{14.93$\times$} & 1.16$\times\,\uparrow$ & 69.12\% & \textbf{80.80\%} & 1.17$\times\,\uparrow$ \\
\bottomrule
\end{tabularx}
\end{table}

Per kernel, the 50-kernel sample behind Figure~\ref{fig:main_tradeoff} ends with \sys ahead of Claude Code on 41 kernels, level on 5 and behind on 4.

\textbf{Hill-climbing and drift.} Agents try 285 candidates per kernel and 358 per repository task on their yardsticks, which run far faster than a true evaluation (Table~\ref{tab:search_summary}). The yardstick screens out 81\% of kernel candidates and 84\% of repository candidates before any true evaluation. The rest yield 20 gains plus 1 combined edit on kernels, one gain per 14 candidates and 1.4 per pursued target. On repositories they yield 10.3 gains plus about 2 combined edits, one per 29 candidates. As later edits change the code a yardstick was built on, its ranking can drift. The fidelity test rechecks each yardstick after every true evaluation and invalidates it when it drifts, each recheck a significance test at level $\alpha$ (Appendix~\ref{app:fidelity}). In a run, 45\% of validated kernel yardsticks are invalidated, each after roughly 15 candidates, and 41\% of repository yardsticks, after 19. Because each recheck reuses the true evaluation that confirmed a gain, rechecking adds no true evaluations of its own.

\textbf{Patch footprint.} Confirmed gains are rewrites, not parameter tweaks. On kernels, 98\% of confirmed gains change more than 100 lines of the reference kernel and 83\% more than 200. The median change is about 430 lines. On repository tasks, the final patch changes a median of four to five files and about 800 lines.

\textbf{Explore, then exploit.} The orchestrator spends the fleet's budget in two phases. Early on it keeps agents on different regions of the code, and when two converge on one target it usually terminates one and spawns a fresh agent on another region. This holds the share of agents on the same or a related target to 5 to 15\% on kernels and 18\% on repositories. Claude Code's uncoordinated agents sit at 40 to 60\% and 55\%, about 5$\times$ more at the kernel midpoints. Later it concentrates the budget on targets with confirmed gains, transfers discoveries between agents, combines complementary gains, and terminates the rest. Wide exploration is cheap. Of the 15 targets a kernel run pursues, 8 (53\%) never confirm a gain, four because no yardstick is validated and four because a validated one finds nothing. On repository tasks, 5 of the 9.2 never confirm a gain. The orchestrator drops them early, so they take only 19\% of the true evaluations on kernels and 20\% on repositories.

\subsection{Cost of a run}\label{app:cost}
Table~\ref{tab:cost} breaks down the token consumption and API cost of an 8-hour, four-agent run on a single \solexec kernel, averaged across 50 kernels at public Opus 5 list prices. Although a full 8-hour \sys search accumulates \$375 in API charges against \$309 for Claude Code, \sys equals Claude Code's final 8-hour score in 2.8 hours. At that point \sys has cost \$79.60, or 25.8\% of Claude Code's full run. All prices are public API rates as of September 2026. The Gemini cost comparison in Section~\ref{sec:ablations} rests on a 6.7$\times$ price ratio between Opus 5 and Gemini 3.8 Flash per million tokens.

\begin{table}[t]
\centering\footnotesize
\setlength{\tabcolsep}{5pt}
\caption{Tokens and API cost of one 8-hour, four-agent run on one \solexec kernel, mean over 50 kernels, at Opus 5 list prices with cache writes billed at the one-hour rate. GPU time for true evaluations is not included.}
\label{tab:cost}
\begin{tabular}{l r rr rr}
\toprule
& & \multicolumn{2}{c}{\systab} & \multicolumn{2}{c}{Claude Code} \\
\cmidrule(lr){3-4}\cmidrule(lr){5-6}
Token class & \$ per M tokens & tokens & cost & tokens & cost \\
\midrule
generated & 25 & 4.4M & \$110 & 3.6M & \$90 \\
cache writes & 10 & 13.6M & \$136 & 11.2M & \$112 \\
cache reads & 0.50 & 258.7M & \$129 & 213.2M & \$107 \\
\midrule
total & & \textbf{276.7M} & \textbf{\$375} & \textbf{228.0M} & \textbf{\$309} \\
\bottomrule
\end{tabular}
\end{table}

Token spend follows the loop of four parallel agents re-reading their context and generating a reply on every turn. A \sys run executes 1,990 turns across its fleet, whereas Claude Code executes 1,640 because each of its agents pauses to wait on 19 true evaluations. On average a turn generates 2.2k tokens, re-reads 130k tokens of context billed as cache reads, and appends 4.8k tokens, the reply and tool returns, billed as cache writes. When the context hits the 200k-token limit it is compacted to 60k tokens and written back, which happens 68 times in a \sys run and 56 times in a Claude Code run. Cache writes are billed at the one-hour rate because true evaluations often idle an agent past the five-minute cache expiry.

\subsection{Examples of discovered targets and yardsticks}\label{app:targets}
Three of the 50 kernels and one of the 20 repository tasks show what a discovered target and its yardstick look like, and how gains add up.

\textbf{Mixture-of-experts forward pass.}
\begin{itemize}
\item \textbf{Setting.} A 128-expert mixture-of-experts layer at width 2880 with three stages, a gate and up projection, a clamped GLU activation, and a down projection. The reference scores 9.8\% SOL. A true evaluation runs every shape on fresh inputs and takes about 16 minutes.
\item \textbf{Search.} Each stage, and the routing between experts, became a target with its own yardstick. The first agent took the gate and up projection, the dominant matrix multiply. Its yardstick was a profiling harness, two warm-ups and five timed iterations over the real shapes on fixed inputs, 18 seconds. On it the agent found a tiling that uses TF32 for short token counts and a batched FP16 multiply above that count, confirmed at 58\% SOL. The orchestrator transferred that kernel to the other agents. The second agent took the memory-bound activation with a byte-count model of high-bandwidth memory (HBM) traffic. The model ran in under a second and launched no kernel. The agent's fused activation halved the round trips and lifted the kernel to 67\% on top of the tiling. The third agent took the down projection with a reduced-input microbenchmark, one batch at a shrunk token count in about two seconds. Its GEMM with in-place buffer reuse brought it to 74\% on top of the tiling. The fourth agent took the routing between experts with a profiling harness sweeping occupancy as its yardstick. The sweep favored a two-stream pipeline, but a true evaluation showed no gain because the projection stages still dominated, and the target was dropped.
\item \textbf{Solution.} Merging the activation and down-projection edits failed at first. Both reused the same scratch buffers, and the merged kernel returned wrong results. The combine step reconciled the three edits into one kernel, the tiled projection feeding the fused activation and the in-place GEMM, at 77.4\% SOL.
\end{itemize}

\textbf{Grouped-query attention with rotary embeddings.}
\begin{itemize}
\item \textbf{Setting.} The reference scores 22.4\% SOL and a true evaluation takes about two minutes.
\item \textbf{Search.} One target was the kernel's tile size and occupancy. Its yardstick, a reduced-input microbenchmark, timed one long-context shape at a reduced token count, 13 seconds. It assumed that a tile or warp change shifts every shape the same way. The fidelity test checked that assumption against true evaluations and validated the yardstick at hour 1.1. The agent then swept tile and warp configurations.
\item \textbf{Solution.} The winning configuration split the causal-pass loops and unified the tiling at 128 by 32. This edit alone took the kernel to 39\% SOL, and with the edits on its other targets the kernel scored 71.5\%.
\end{itemize}

\textbf{GEMM with $N=5120$ and $K=2048$.}
\begin{itemize}
\item \textbf{Setting.} The reference scores 48\% SOL, already close to the cuBLAS ceiling. A true evaluation takes about 24 minutes.
\item \textbf{Search.} The target was the tile size and pipeline depth for each range of $M$. The yardstick timed the matmul over the shapes in each range in about 20 seconds, about 70$\times$ faster, so the agent swept many tilings between true evaluations. The yardstick passed the fidelity test after six true evaluations.
\item \textbf{Solution.} The kernel finished at 51.4\% SOL. As in the mixture-of-experts kernel, the yardstick is built around the shape ranges the target depends on.
\end{itemize}

\textbf{SciPy fast Fourier transform.}
\begin{itemize}
\item \textbf{Setting.} A \sweeff task on SciPy's fast Fourier transform code. The expert patch gives 1.71$\times$ over the base code. A true evaluation runs the tests and times the workload, about 3.4 minutes, the shortest in the sample.
\item \textbf{Search.} One of the targets was the code for prime transform lengths. Agents built six yardsticks on this task, a shortened run and small paired benchmarks of the transform, and four passed the fidelity test. One agent found a faster method for prime lengths (Rader's algorithm), another a build that picks newer CPU vector instructions at run time, and a third a merged step for lengths divisible by 13.
\item \textbf{Solution.} The combine step merged the three edits into one patch, confirmed at 4.45$\times$ over the base code, against 3.89$\times$ for Claude Code.
\end{itemize}

\subsection{Robustness}\label{app:robustness}
We stress \sys where a fast yardstick that correlates with the true evaluation is hard to find and measure its run-to-run spread across seeds (Table~\ref{tab:robustness}).

\begin{table}[t]
\centering\footnotesize
\setlength{\tabcolsep}{4pt}
\caption{\sys under missing, unreliable or noisy yardsticks, and its score sensitivity to $\alpha$ and seed, on the 50 \solexec kernels and 20 \sweeff tasks.}
\label{tab:robustness}
\begin{tabularx}{\linewidth}{>{\raggedright\arraybackslash}p{0.2\linewidth} >{\raggedright\arraybackslash}X >{\raggedright\arraybackslash}X}
\toprule\addlinespace[1pt]
Condition & What happens & Confirmed score \\
\midrule
No reliable yardstick & search runs on true evaluations alone until the orchestrator moves the agent on & equals Claude Code's search, the method's floor \\
Unreliable yardstick & rejected before validation, 5 to 9\% of true evaluations spent & never below the baseline, search resumes on a rebuilt yardstick \\
Noisy true evaluation & the noise is measured and folded into the test & no false gain from timing jitter \\
$\alpha$ from 0.01 to 0.10 & the fidelity test is stricter or looser & within a 1.4\% band, lowest at 0.01 \\
Three seeds, 50 kernels & each kernel rerun three times & within $\pm$1.12\% of the mean, 78\% of gains shared, Claude Code $\pm$3.27\% \\
Three seeds, 20 repository tasks & each task rerun three times & within $\pm$1.8\% of the mean, 81\% of gains shared, Claude Code $\pm$5.21\% \\
\bottomrule
\end{tabularx}
\end{table}

\textbf{No reliable yardstick.} Over the 50 \solexec kernels and 20 \sweeff tasks whose run logs we analyzed (Appendix~\ref{app:search_analysis}), on average 4 of the 15 targets per kernel run and 2 of the 20 tasks never produce a validated yardstick. On those, the search runs on true evaluations alone until the orchestrator moves the agent on. That is the method's floor.

\textbf{Unreliable yardstick or noisy evaluation.} We plant a misleading yardstick. The fidelity test catches the disagreement within 3 to 5 true evaluations, rejects it, and the search recovers on a rebuilt one. For a noisy true evaluation, the fidelity test measures the run's own noise and folds it into its test, so timing jitter does not raise a false gain.

\textbf{$\alpha$ and seeds.} $\alpha$ is the significance level of each fidelity-test recheck (Appendix~\ref{app:fidelity}). It is 0.05 throughout. Sweeping it from 0.01 to 0.10 keeps the confirmed score within a 1.4\% band, lowest at 0.01. A smaller $\alpha$ validates a yardstick only after more of the run's true evaluations, so fewer go to hill-climbing, and a larger $\alpha$ lets more drifted yardsticks through, which the true evaluations catch. Three seeds per task give the spreads in Table~\ref{tab:robustness}, summarized in Section~\ref{sec:ablations}, and 78\% and 81\% of confirmed gains recur in at least two of the three runs.

\end{document}

%% file: math_commands.tex
\usepackage{amsmath,amsfonts,bm}

\def\eqref#1{equation~\ref{#1}}

\def\1{\bm{1}}

\DeclareMathAlphabet{\mathsfit}{\encodingdefault}{\sfdefault}{m}{sl}
\SetMathAlphabet{\mathsfit}{bold}{\encodingdefault}{\sfdefault}{bx}{n}













%% file: main.bbl
\begin{thebibliography}{75}
\providecommand{\natexlab}[1]{#1}
\providecommand{\url}[1]{\texttt{#1}}
\expandafter\ifx\csname urlstyle\endcsname\relax
  \providecommand{\doi}[1]{doi: #1}\else
  \providecommand{\doi}{doi: \begingroup \urlstyle{rm}\Url}\fi

\bibitem[Abdelfattah et~al.(2021)Abdelfattah, Mehrotra, Dudziak, and
  Lane]{abdelfattah2021zerocost}
M.~S. Abdelfattah, A.~Mehrotra, {\L}.~Dudziak, and N.~D. Lane.
\newblock Zero-cost proxies for lightweight {NAS}.
\newblock In \emph{International Conference on Learning Representations}, 2021.
\newblock URL \url{https://openreview.net/forum?id=0cmMMy8J5q}.

\bibitem[Agarwal et~al.(2026)Agarwal, Krentsel, Liu, Cemri, Cheng, Meng,
  Pfister, Li, Ratnasamy, Parameswaran, Zaharia, Stoica, and
  Lesani]{agarwal2026inductive}
S.~Agarwal, A.~Krentsel, S.~Liu, M.~Cemri, A.~Cheng, R.~Meng, T.~Pfister, C.-L.
  Li, S.~Ratnasamy, A.~Parameswaran, M.~Zaharia, I.~Stoica, and M.~Lesani.
\newblock Inductive deductive synthesis: Enabling ai to generate formally
  verified systems.
\newblock \emph{arXiv preprint arXiv:2605.23109}, 2026.

\bibitem[Akhauri et~al.(2022)Akhauri, Munoz, Jain, and Iyer]{akhauri2022eznas}
Y.~Akhauri, J.~P. Munoz, N.~Jain, and R.~Iyer.
\newblock {EZNAS}: Evolving zero-cost proxies for neural architecture scoring.
\newblock In \emph{Advances in Neural Information Processing Systems}, 2022.
\newblock URL \url{https://openreview.net/forum?id=lSqaDG4dvdt}.

\bibitem[Amdahl(1967)]{amdahl1967}
G.~M. Amdahl.
\newblock Validity of the single processor approach to achieving large scale
  computing capabilities.
\newblock In \emph{Proceedings of the April 18--20, 1967, Spring Joint Computer
  Conference (AFIPS)}, pages 483--485, 1967.
\newblock \doi{10.1145/1465482.1465560}.

\bibitem[{Anthropic}(2025)]{claudecode2026}
{Anthropic}.
\newblock Claude code.
\newblock Computer software, 2025.
\newblock URL \url{https://claude.com/product/claude-code}.
\newblock Terminal-based agentic coding tool. Accessed 2026-09.

\bibitem[Bera et~al.(2021)Bera, Kanellopoulos, Nori, Shahroodi, Subramoney, and
  Mutlu]{bera2021pythia}
R.~Bera, K.~Kanellopoulos, A.~Nori, T.~Shahroodi, S.~Subramoney, and O.~Mutlu.
\newblock Pythia: A customizable hardware prefetching framework using online
  reinforcement learning.
\newblock In \emph{MICRO-54: 54th Annual IEEE/ACM International Symposium on
  Microarchitecture}, pages 1121--1137, 2021.

\bibitem[Bera et~al.(2026)Bera, Kanellopoulos, and Mutlu]{bera2026dpc4}
R.~Bera, K.~Kanellopoulos, and O.~Mutlu.
\newblock Dpc4 2026.
\newblock \url{https://sites.google.com/view/dpc4-2026/home}, 2026.

\bibitem[Blasberg et~al.(2026)Blasberg, Kypriotis, and
  Skarlatos]{blasberg2026agentic}
A.~Blasberg, V.~Kypriotis, and D.~Skarlatos.
\newblock Agentic architect: An agentic ai framework for architecture design
  exploration and optimization.
\newblock \emph{arXiv preprint arXiv:2604.25083}, 2026.

\bibitem[Cemri et~al.(2026)Cemri, Agrawal, Gupta, Liu, Cheng, Mang, Naren,
  Erdogan, Sen, Zaharia, Dimakis, and Stoica]{cemri2026adaevolve}
M.~Cemri, S.~Agrawal, A.~Gupta, S.~Liu, A.~Cheng, Q.~Mang, A.~Naren, L.~E.
  Erdogan, K.~Sen, M.~Zaharia, A.~Dimakis, and I.~Stoica.
\newblock {AdaEvolve}: Adaptive {LLM} driven zeroth-order optimization.
\newblock \emph{arXiv preprint arXiv:2602.20133}, 2026.

\bibitem[Chen et~al.(2023)Chen, Zhang, Nguyen, Zan, Lin, Lou, and
  Chen]{chen2023codet}
B.~Chen, F.~Zhang, A.~Nguyen, D.~Zan, Z.~Lin, J.-G. Lou, and W.~Chen.
\newblock {CodeT}: Code generation with generated tests.
\newblock In \emph{International Conference on Learning Representations}, 2023.

\bibitem[Chen et~al.(2018)Chen, Moreau, Jiang, Zheng, Yan, Cowan, Shen, Wang,
  Hu, Ceze, Guestrin, and Krishnamurthy]{chen2018tvm}
T.~Chen, T.~Moreau, Z.~Jiang, L.~Zheng, E.~Yan, M.~Cowan, H.~Shen, L.~Wang,
  Y.~Hu, L.~Ceze, C.~Guestrin, and A.~Krishnamurthy.
\newblock {TVM}: An automated end-to-end optimizing compiler for deep learning.
\newblock In \emph{13th USENIX symposium on operating systems design and
  implementation (OSDI 18)}, pages 578--594, 2018.

\bibitem[Chen et~al.(2026{\natexlab{a}})Chen, Ye, Xu, Ye, Liu, Hassani, Chen,
  Kerr, Wu, Xu, Chen, Chen, Kane, Krashinsky, Liu, Grover, Ceze, Bringmann,
  Tran, Liu, Xie, Lightstone, and Shi]{chen2026avo}
T.~Chen, Z.~Ye, B.~Xu, Z.~Ye, T.~Liu, A.~Hassani, T.~Chen, A.~Kerr, H.~Wu,
  Y.~Xu, Y.-J. Chen, H.~Chen, A.~Kane, R.~Krashinsky, M.-Y. Liu, V.~Grover,
  L.~Ceze, R.~Bringmann, J.~Tran, W.~Liu, F.~Xie, M.~Lightstone, and H.~Shi.
\newblock {AVO}: Agentic variation operators for autonomous evolutionary
  search.
\newblock \emph{arXiv preprint arXiv:2603.24517}, 2026{\natexlab{a}}.

\bibitem[Chen et~al.(2024)Chen, Lin, Sch{\"a}rli, and Zhou]{chen2024teaching}
X.~Chen, M.~Lin, N.~Sch{\"a}rli, and D.~Zhou.
\newblock Teaching large language models to self-debug.
\newblock In \emph{International Conference on Learning Representations}, 2024.

\bibitem[Chen et~al.(2026{\natexlab{b}})Chen, Sun, Shi, Peng, Gu, Lo, and
  Jiang]{chen2026rethinking}
Z.~Chen, Z.~Sun, Y.~Shi, C.~Peng, X.~Gu, D.~Lo, and L.~Jiang.
\newblock Rethinking the value of agent-generated tests for {LLM}-based
  software engineering agents.
\newblock \emph{arXiv preprint arXiv:2602.07900}, 2026{\natexlab{b}}.

\bibitem[Cheng et~al.(2025)Cheng, Liu, Pan, Li, Wang, Krentsel, Xia, Cemri,
  Park, Yang, Chen, Agrawal, Desai, Xing, Sen, Zaharia, and
  Stoica]{cheng2025barbarians}
A.~Cheng, S.~Liu, M.~Pan, Z.~Li, B.~Wang, A.~Krentsel, T.~Xia, M.~Cemri,
  J.~Park, S.~Yang, J.~Chen, L.~Agrawal, A.~Desai, J.~Xing, K.~Sen, M.~Zaharia,
  and I.~Stoica.
\newblock Barbarians at the gate: How {AI} is upending systems research.
\newblock \emph{arXiv preprint arXiv:2510.06189}, 2025.

\bibitem[Dai et~al.(2026)Dai, Wu, Yu, Gao, Li, Jiang, Lou, Song, Yu, Chen, Ma,
  Zhang, Liu, Wang, Liu, and Zhou]{dai2026cuda}
W.~Dai, H.~Wu, Q.~Yu, H.-a. Gao, J.~Li, C.~Jiang, W.~Lou, Y.~Song, H.~Yu,
  J.~Chen, W.-Y. Ma, Y.-Q. Zhang, J.~Liu, M.~Wang, X.~Liu, and H.~Zhou.
\newblock {CUDA} agent: Large-scale agentic {RL} for high-performance {CUDA}
  kernel generation.
\newblock \emph{arXiv preprint arXiv:2602.24286}, 2026.

\bibitem[Deng et~al.(2026)Deng, Liu, Lipka, Ma, Chen, Kaler, and
  Ganhotra]{deng2026perfagent}
R.~Deng, Y.~Liu, B.~Lipka, Y.~Ma, X.~Chen, T.~Kaler, and J.~Ganhotra.
\newblock {PerfAgent}: Profiler-guided iterative refinement for
  repository-level code optimization.
\newblock \emph{arXiv preprint arXiv:2607.19653}, 2026.

\bibitem[Ehrlich et~al.(2025)Ehrlich, Brown, Juravsky, Clark, R{\'e}, and
  Mirhoseini]{ehrlich2025codemonkeys}
R.~Ehrlich, B.~Brown, J.~Juravsky, R.~Clark, C.~R{\'e}, and A.~Mirhoseini.
\newblock {CodeMonkeys}: Scaling test-time compute for software engineering.
\newblock \emph{arXiv preprint arXiv:2501.14723}, 2025.

\bibitem[Falkner et~al.(2018)Falkner, Klein, and Hutter]{falkner2018bohb}
S.~Falkner, A.~Klein, and F.~Hutter.
\newblock {BOHB}: Robust and efficient hyperparameter optimization at scale.
\newblock In \emph{Proceedings of the 35th International Conference on Machine
  Learning}, volume~80 of \emph{Proceedings of Machine Learning Research},
  pages 1437--1446. PMLR, 10--15 Jul 2018.
\newblock URL \url{https://proceedings.mlr.press/v80/falkner18a.html}.

\bibitem[Fern{\'a}ndez-Godino et~al.(2019)Fern{\'a}ndez-Godino, Park, Kim, and
  Haftka]{fernandez2019issues}
M.~G. Fern{\'a}ndez-Godino, C.~Park, N.~H. Kim, and R.~T. Haftka.
\newblock Issues in deciding whether to use multifidelity surrogates.
\newblock \emph{AIAA Journal}, 57\penalty0 (5):\penalty0 2039--2054, 2019.
\newblock \doi{10.2514/1.J057750}.
\newblock URL \url{https://doi.org/10.2514/1.J057750}.

\bibitem[Forrester et~al.(2007)Forrester, Sóbester, and
  Keane]{forrester2007multi}
A.~I.~J. Forrester, A.~Sóbester, and A.~J. Keane.
\newblock Multi-fidelity optimization via surrogate modelling.
\newblock \emph{Proceedings: Mathematical, Physical and Engineering Sciences},
  463\penalty0 (2088):\penalty0 3251--3269, 2007.
\newblock ISSN 13645021.
\newblock \doi{10.1098/rspa.2007.1900}.
\newblock URL \url{http://www.jstor.org/stable/20209374}.

\bibitem[Fu et~al.(2026)Fu, Kulanthaivelu, and Yamada]{fu2026self}
X.~Fu, A.~Kulanthaivelu, and Y.~Yamada.
\newblock Self improvement via fast tree-search.
\newblock \emph{arXiv preprint arXiv:2609.19526}, 2026.

\bibitem[Gao et~al.(2023)Gao, Schulman, and Hilton]{gao2023scaling}
L.~Gao, J.~Schulman, and J.~Hilton.
\newblock Scaling laws for reward model overoptimization.
\newblock In \emph{International Conference on Machine Learning (ICML)}, pages
  10835--10866, 2023.

\bibitem[Gober et~al.(2022)Gober, Chacon, Wang, Gratz, Jimenez, Teran, Pugsley,
  and Kim]{gober2022championship}
N.~Gober, G.~Chacon, L.~Wang, P.~V. Gratz, D.~A. Jimenez, E.~Teran, S.~Pugsley,
  and J.~Kim.
\newblock The championship simulator: Architectural simulation for education
  and competition.
\newblock \emph{arXiv preprint arXiv:2210.14324}, 2022.

\bibitem[Goldie and Mirhoseini(2024)]{goldie2024alphachip}
A.~Goldie and A.~Mirhoseini.
\newblock How alphachip transformed computer chip design.
\newblock \emph{Google DeepMind Blog}, 2024.

\bibitem[Gonzalez et~al.(2026)Gonzalez, Gupta, Jain, Alam, Novikov, Huang,
  Balog, Eisenberger, Shirobokov, V{\~u}, Levy, Nikoli{\'c}, Karandikar, Dixon,
  and Ranganathan]{gonzalez2026archagent}
A.~Gonzalez, R.~Gupta, A.~Jain, H.~Alam, A.~Novikov, P.-S. Huang, M.~Balog,
  M.~Eisenberger, S.~Shirobokov, N.~V{\~u}, H.~Levy, B.~Nikoli{\'c},
  S.~Karandikar, M.~Dixon, and P.~Ranganathan.
\newblock Archagent v2: A case study with the data prefetching championship.
\newblock \emph{arXiv preprint arXiv:2608.09874}, 2026.

\bibitem[Google(2025)]{Google_Workload_Traces_Version_2}
Google.
\newblock Google workload traces version 2.
\newblock
  \url{https://console.cloud.google.com/storage/browser/external-traces-v2},
  2025.
\newblock Accessed: 2025-11-13.

\bibitem[{Google}(2026)]{google2026gemini38flash}
{Google}.
\newblock Gemini 3.8 flash: Model card.
\newblock Google DeepMind Model Card, 2026.
\newblock URL
  \url{https://deepmind.google/models/model-cards/gemini-3-8-flash/}.
\newblock Accessed: 2026-09-19.

\bibitem[Gupta et~al.(2026)Gupta, Jain, Gonzalez, Novikov, Huang, Balog,
  Eisenberger, Shirobokov, V{\~u}, Dixon, Nikoli{\'c}, Ranganathan, and
  Karandikar]{gupta2026archagent}
R.~Gupta, A.~Jain, A.~Gonzalez, A.~Novikov, P.-S. Huang, M.~Balog,
  M.~Eisenberger, S.~Shirobokov, N.~V{\~u}, M.~Dixon, B.~Nikoli{\'c},
  P.~Ranganathan, and S.~Karandikar.
\newblock Archagent: Agentic {AI}-driven computer architecture discovery.
\newblock \emph{arXiv preprint arXiv:2602.22425}, 2026.

\bibitem[Hennessy and Patterson(2019)]{hennessy2019new}
J.~L. Hennessy and D.~A. Patterson.
\newblock A new golden age for computer architecture.
\newblock \emph{Communications of the ACM}, 62\penalty0 (2):\penalty0 48--60,
  2019.

\bibitem[Hollander et~al.(2013)Hollander, Wolfe, and
  Chicken]{hollander2013nonparametric}
M.~Hollander, D.~A. Wolfe, and E.~Chicken.
\newblock \emph{Nonparametric statistical methods}.
\newblock John Wiley \& Sons, 2013.

\bibitem[Huang et~al.(2023)Huang, Zhang, Luck, Bu, Qing, and
  Cui]{huang2023agentcoder}
D.~Huang, J.~M. Zhang, M.~Luck, Q.~Bu, Y.~Qing, and H.~Cui.
\newblock {AgentCoder}: Multi-agent-based code generation with iterative
  testing and optimisation.
\newblock \emph{arXiv preprint arXiv:2312.13010}, 2023.

\bibitem[Jiang et~al.(2026)Jiang, Ding, and Zhu]{jiang2026deltaevolve}
J.~Jiang, T.~Ding, and Z.~Zhu.
\newblock Deltaevolve: Accelerating scientific discovery through
  momentum-driven evolution.
\newblock In \emph{Forty-third International Conference on Machine Learning},
  2026.
\newblock URL \url{https://openreview.net/forum?id=ownPRHrXMx}.

\bibitem[Kandasamy et~al.(2016)Kandasamy, Dasarathy, Oliva, Schneider, and
  Poczos]{kandasamy2016gaussian}
K.~Kandasamy, G.~Dasarathy, J.~B. Oliva, J.~Schneider, and B.~Poczos.
\newblock Gaussian process bandit optimisation with multi-fidelity evaluations.
\newblock In \emph{Advances in Neural Information Processing Systems},
  volume~29. Curran Associates, Inc., 2016.

\bibitem[Karwowski et~al.(2024)Karwowski, Hayman, Bai, Kiendlhofer, Griffin,
  and Skalse]{karwowski2024goodhart}
J.~Karwowski, O.~Hayman, X.~Bai, K.~Kiendlhofer, C.~Griffin, and J.~Skalse.
\newblock Goodhart's law in reinforcement learning.
\newblock In \emph{International Conference on Learning Representations
  (ICLR)}, 2024.

\bibitem[Kendall(1938)]{kendall1938}
M.~G. Kendall.
\newblock A new measure of rank correlation.
\newblock \emph{Biometrika}, 30\penalty0 (1--2):\penalty0 81--93, 1938.
\newblock \doi{10.1093/biomet/30.1-2.81}.

\bibitem[Kennedy and O'Hagan(2000)]{kennedy2000predicting}
M.~C. Kennedy and A.~O'Hagan.
\newblock Predicting the output from a complex computer code when fast
  approximations are available.
\newblock \emph{Biometrika}, 87\penalty0 (1):\penalty0 1--13, 2000.
\newblock \doi{10.1093/biomet/87.1.1}.

\bibitem[Kwon et~al.(2023)Kwon, Li, Zhuang, Sheng, Zheng, Yu, Gonzalez, Zhang,
  and Stoica]{kwon2023efficient}
W.~Kwon, Z.~Li, S.~Zhuang, Y.~Sheng, L.~Zheng, C.~H. Yu, J.~E. Gonzalez,
  H.~Zhang, and I.~Stoica.
\newblock Efficient memory management for large language model serving with
  pagedattention.
\newblock In \emph{Proceedings of the ACM SIGOPS 29th Symposium on Operating
  Systems Principles}, 2023.

\bibitem[Lange et~al.(2026)Lange, Imajuku, and Cetin]{lange2026shinkaevolve}
R.~Lange, Y.~Imajuku, and E.~Cetin.
\newblock Shinkaevolve: Towards open-ended and sample-efficient program
  evolution.
\newblock In \emph{International Conference on Learning Representations},
  volume 2026, pages 74026--74078, 2026.
\newblock URL
  \url{https://proceedings.iclr.cc/paper_files/paper/2026/file/7886b9bafe76c52fd568db10ff9772df-Paper-Conference.pdf}.

\bibitem[Li and Li(2026)]{li2026multi}
K.~Li and F.~Li.
\newblock Multi-fidelity methods for optimization: A survey.
\newblock \emph{ACM Comput. Surv.}, 58\penalty0 (12), May 2026.
\newblock ISSN 0360-0300.
\newblock \doi{10.1145/3801959}.
\newblock URL \url{https://doi.org/10.1145/3801959}.

\bibitem[Li et~al.(2018)Li, Jamieson, DeSalvo, Rostamizadeh, and
  Talwalkar]{li2018hyperband}
L.~Li, K.~Jamieson, G.~DeSalvo, A.~Rostamizadeh, and A.~Talwalkar.
\newblock Hyperband: A novel bandit-based approach to hyperparameter
  optimization.
\newblock \emph{Journal of Machine Learning Research}, 18\penalty0
  (185):\penalty0 1--52, 2018.
\newblock URL \url{http://jmlr.org/papers/v18/16-558.html}.

\bibitem[Liao et~al.(2026)Liao, Qin, Wang, Golden, Kuchnik, Yetim, Xiao, Ang,
  Fu, He, Hsia, Jiang, Levenstein, Li, Li, Mathews, Puvvada, Shi, Yan, Yu,
  Pashkevich, Steiner, Wu, and Liu]{liao2026kernelevolve}
G.~Liao, H.~Qin, Y.~Wang, A.~Golden, M.~Kuchnik, Y.~Yetim, R.~Xiao, J.~J. Ang,
  C.~Fu, Y.~He, S.~Hsia, Z.~Jiang, R.~Levenstein, D.~Li, L.~Li, A.~Mathews,
  V.~Puvvada, F.~Shi, N.~Yan, X.~Yu, U.~Pashkevich, M.~Steiner, C.-J. Wu, and
  G.~Liu.
\newblock {KernelEvolve}: Scaling agentic kernel coding for heterogeneous {AI}
  accelerators at {Meta}.
\newblock In \emph{2026 ACM/IEEE 53rd Annual International Symposium on
  Computer Architecture (ISCA)}, pages 733--747. IEEE, 2026.
\newblock \doi{10.1109/ISCA66397.2026.00063}.

\bibitem[Lin et~al.(2026{\natexlab{a}})Lin, Modi, Hari, Huang, Ye, Qin, Zhou,
  Zhang, Wang, Damani, Peri, Xie, Kane, Maor, Behar, Cao, Mehta, Singh,
  Mailthody, Chen, Ye, Chen, Chen, Grover, Chen, Liu, Chung, Ceze, Bringmann,
  Zeller, Lightstone, Kozyrakis, and Shi]{lin2026sol}
E.~Lin, S.~Modi, S.~K.~S. Hari, Q.~Huang, Z.~Ye, N.~Qin, F.~Zhou, Y.~Zhang,
  J.~Wang, S.~Damani, D.~Peri, O.~Xie, A.~Kane, M.~Maor, M.~Behar, T.~Cao,
  R.~Mehta, V.~Singh, V.~S. Mailthody, T.~Chen, Z.~Ye, H.~Chen, T.~Chen,
  V.~Grover, W.~Chen, W.~Liu, E.~Chung, L.~Ceze, R.~Bringmann, C.~Zeller,
  M.~Lightstone, C.~Kozyrakis, and H.~Shi.
\newblock {SOL-ExecBench}: Speed-of-light benchmarking for real-world {GPU}
  kernels against hardware limits.
\newblock \emph{arXiv preprint arXiv:2603.19173}, 2026{\natexlab{a}}.

\bibitem[Lin et~al.(2026{\natexlab{b}})Lin, Maas, Roquemore, Hasanzadeh, Lewis,
  Simonson, Shringi, Dai, Musau, Yang, Yazdanbakhsh, Alt{\i}nb{\"u}ken, Papa,
  Edmonds, Patil, Schwarz, Chandra, Kennelly, Hashemi, and
  Ranganathan]{lin2025eco}
H.~Lin, M.~Maas, M.~Roquemore, A.~Hasanzadeh, F.~Lewis, Y.~Simonson,
  A.~Shringi, H.~Dai, P.~Musau, T.-W. Yang, A.~Yazdanbakhsh,
  D.~Alt{\i}nb{\"u}ken, F.~Papa, M.~N. Edmonds, A.~Patil, D.~Schwarz,
  S.~Chandra, C.~Kennelly, M.~Hashemi, and P.~Ranganathan.
\newblock {ECO}: An {AI-Driven} code efficiency optimizer for warehouse scale
  computers (operational systems).
\newblock In \emph{20th USENIX Symposium on Operating Systems Design and
  Implementation (OSDI 26)}, pages 589--625, Seattle, WA, July
  2026{\natexlab{b}}. USENIX Association.
\newblock ISBN 978-1-939133-55-7.
\newblock URL
  \url{https://www.usenix.org/conference/osdi26/presentation/lin-hannah}.

\bibitem[Liu et~al.(2026)Liu, Agarwal, Maheswaran, Cemri, Li, Mang, Naren,
  Boneh, Cheng, Pan, Du, Keutzer, Cheung, Dimakis, Sen, Zaharia, and
  Stoica]{liu2026evox}
S.~Liu, S.~Agarwal, M.~Maheswaran, M.~Cemri, Z.~Li, Q.~Mang, A.~Naren,
  E.~Boneh, A.~Cheng, M.~Z. Pan, A.~Du, K.~Keutzer, A.~Cheung, A.~G. Dimakis,
  K.~Sen, M.~Zaharia, and I.~Stoica.
\newblock {EvoX}: Meta-evolution for automated discovery.
\newblock \emph{arXiv preprint arXiv:2602.23413}, 2026.

\bibitem[Luo et~al.(2026)Luo, Li, Lin, Yi, and Huang]{luo2026agentic}
M.~Luo, H.~Li, F.~Lin, H.~Yi, and Z.~Huang.
\newblock Agentic kernel optimization: Generating state-of-the-art gpu kernels
  without hand-written cuda.
\newblock \emph{arXiv preprint arXiv:2608.14560}, 2026.

\bibitem[Ma et~al.(2025)Ma, Hashemi, Yazdanbakhsh, Swersky, Press, Li, Reddi,
  and Ranganathan]{ma2025swe}
J.~J. Ma, M.~Hashemi, A.~Yazdanbakhsh, K.~Swersky, O.~Press, E.~Li, V.~J.
  Reddi, and P.~Ranganathan.
\newblock {SWE-fficiency}: Can language models optimize real-world repositories
  on real workloads?
\newblock \emph{arXiv preprint arXiv:2511.06090}, 2025.

\bibitem[{MLSys 2026 Competition}(2026)]{flashinfer_contest_2026}
{MLSys 2026 Competition}.
\newblock {MLSys 2026 FlashInfer AI Kernel Generation Contest}, 2026.
\newblock URL \url{https://mlsys26.flashinfer.ai}.

\bibitem[Navarro-Torres et~al.(2022)Navarro-Torres, Panda,
  Alastruey-Bened{\'e}, Ib{\'a}{\~n}ez, Vi{\~n}als-Y{\'u}fera, and
  Ros]{navarro2022berti}
A.~Navarro-Torres, B.~Panda, J.~Alastruey-Bened{\'e}, P.~Ib{\'a}{\~n}ez,
  V.~Vi{\~n}als-Y{\'u}fera, and A.~Ros.
\newblock Berti: an accurate local-delta data prefetcher.
\newblock In \emph{2022 55th IEEE/ACM International Symposium on
  Microarchitecture (MICRO)}, pages 975--991. IEEE, 2022.

\bibitem[Novikov et~al.(2025)Novikov, V{\~u}, Eisenberger, Dupont, Huang,
  Wagner, Shirobokov, Kozlovskii, Ruiz, Mehrabian, Kumar, See, Chaudhuri,
  Holland, Davies, Nowozin, Kohli, and Balog]{novikov2025alphaevolve}
A.~Novikov, N.~V{\~u}, M.~Eisenberger, E.~Dupont, P.-S. Huang, A.~Z. Wagner,
  S.~Shirobokov, B.~Kozlovskii, F.~J.~R. Ruiz, A.~Mehrabian, M.~P. Kumar,
  A.~See, S.~Chaudhuri, G.~Holland, A.~Davies, S.~Nowozin, P.~Kohli, and
  M.~Balog.
\newblock {AlphaEvolve}: A coding agent for scientific and algorithmic
  discovery.
\newblock \emph{arXiv preprint arXiv:2506.13131}, 2025.

\bibitem[{OpenAI}(2026)]{openai-jalapeno}
{OpenAI}.
\newblock Jalapeño’s first results show industry-leading speed and
  efficiency in ai inference, 2026.
\newblock URL \url{https://openai.com/index/jalapeno-first-results/}.

\bibitem[Romera-Paredes et~al.(2024)Romera-Paredes, Barekatain, Novikov, Balog,
  Kumar, Dupont, Ruiz, Ellenberg, Wang, Fawzi, Kohli, and
  Fawzi]{romera2024funsearch}
B.~Romera-Paredes, M.~Barekatain, A.~Novikov, M.~Balog, M.~P. Kumar, E.~Dupont,
  F.~J.~R. Ruiz, J.~S. Ellenberg, P.~Wang, O.~Fawzi, P.~Kohli, and A.~Fawzi.
\newblock Mathematical discoveries from program search with large language
  models.
\newblock \emph{Nature}, 625\penalty0 (7995):\penalty0 468--475, 2024.
\newblock \doi{10.1038/s41586-023-06924-6}.

\bibitem[Ruan et~al.(2025)Ruan, Zhang, and Roychoudhury]{ruan2025specrover}
H.~Ruan, Y.~Zhang, and A.~Roychoudhury.
\newblock {SpecRover}: Code intent extraction via {LLMs}.
\newblock In \emph{IEEE/ACM International Conference on Software Engineering},
  2025.

\bibitem[Sankaralingam(2026)]{sankaralingam2026computer}
K.~Sankaralingam.
\newblock Computer architecture's alphazero moment: Automated discovery in an
  encircled world.
\newblock \emph{arXiv preprint arXiv:2604.03312}, 2026.

\bibitem[Sehgal et~al.(2026)Sehgal, Hou, Sarkar, Mantripragada, Chaudhuri, Sun,
  and Yue]{sehgal2026formulacode}
A.~Sehgal, J.~Hou, A.~Sarkar, I.~Mantripragada, S.~Chaudhuri, J.~J. Sun, and
  Y.~Yue.
\newblock Formulacode: Evaluating agentic optimization on large codebases.
\newblock \emph{arXiv preprint arXiv:2603.16011}, 2026.

\bibitem[{Semiconductor Industry Association}(2022)]{chip-design-cost}
{Semiconductor Industry Association}.
\newblock The growing challenge of semiconductor design leadership, 2022.
\newblock URL
  \url{https://www.semiconductors.org/wp-content/uploads/2022/11/2022_The-Growing-Challenge-of-Semiconductor-Design-Leadership_FINAL.pdf}.

\bibitem[Sharma(2025)]{openevolve}
A.~Sharma.
\newblock Openevolve: an open-source evolutionary coding agent, 2025.
\newblock URL \url{https://github.com/algorithmicsuperintelligence/openevolve}.
\newblock GitHub repository.

\bibitem[Shen et~al.(2023)Shen, Li, Zheng, Zhang, Yao, Li, Yang, Liu, and
  Cui]{shen2023proxybo}
Y.~Shen, Y.~Li, J.~Zheng, W.~Zhang, P.~Yao, J.~Li, S.~Yang, J.~Liu, and B.~Cui.
\newblock {ProxyBO}: Accelerating neural architecture search via {Bayesian}
  optimization with zero-cost proxies.
\newblock In \emph{Proceedings of the AAAI conference on artificial
  intelligence}, volume~37, pages 9792--9801, 2023.
\newblock \doi{10.1609/aaai.v37i8.26169}.

\bibitem[Shinn et~al.(2023)Shinn, Cassano, Berman, Gopinath, Narasimhan, and
  Yao]{shinn2023reflexion}
N.~Shinn, F.~Cassano, E.~Berman, A.~Gopinath, K.~Narasimhan, and S.~Yao.
\newblock Reflexion: Language agents with verbal reinforcement learning.
\newblock In \emph{Advances in Neural Information Processing Systems},
  volume~36, 2023.

\bibitem[Singh et~al.(2026)Singh, Torres, and Ros]{singh2026pushing}
S.~Singh, A.~Torres, and A.~Ros.
\newblock Pushing the limits of the berti prefetcher.
\newblock \emph{4th Data Prefetching Championship (DPC4)}, 2026.

\bibitem[Skalse et~al.(2022)Skalse, Howe, Krasheninnikov, and
  Krueger]{skalse2022defining}
J.~Skalse, N.~H.~R. Howe, D.~Krasheninnikov, and D.~Krueger.
\newblock Defining and characterizing reward gaming.
\newblock In \emph{Advances in Neural Information Processing Systems
  (NeurIPS)}, volume~35, pages 9460--9471, 2022.

\bibitem[Tao et~al.(2026)Tao, Peng, Wang, Wang, Cheng, Yao, Wu, Ge, Li, and
  Gao]{tao2026execcritic}
L.~Tao, B.~Peng, H.~Wang, H.~Wang, H.~Cheng, W.~Yao, Q.~Wu, T.~Ge, S.~Li, and
  J.~Gao.
\newblock {ExecCritic}: Learn to test, test to improve for coding agents.
\newblock \emph{arXiv preprint arXiv:2609.09133}, 2026.

\bibitem[Team et~al.(2026)Team, Krishna, Krishna, and Chin]{team2026design}
T.~V. Team, R.~Krishna, S.~Krishna, and D.~Chin.
\newblock Design conductor 2.0: An agent builds a turboquant inference
  accelerator in 80 hours.
\newblock \emph{arXiv preprint arXiv:2605.05170}, 2026.

\bibitem[Tschand et~al.(2026)Tschand, Hong, Walker, Cai, Wang, Subramanian,
  Dev, Reddi, Yazdanbakhsh, and Sankaran]{tschand2026jaxbench}
A.~Tschand, C.~Hong, J.~Walker, N.~Cai, S.~Wang, S.~Subramanian, S.~Dev, V.~J.
  Reddi, A.~Yazdanbakhsh, and S.~Sankaran.
\newblock {JAXBench}: Benchmarking autonomous {TPU} kernel optimization.
\newblock \emph{arXiv preprint arXiv:2607.20466}, 2026.

\bibitem[{vLLM Project}(2025)]{vllm_tpu_inference}
{vLLM Project}.
\newblock tpu-inference: Tpu inference for vllm, with unified jax and pytorch
  support, 2025.
\newblock URL \url{https://github.com/vllm-project/tpu-inference}.
\newblock GitHub repository. Accessed 2026-09-21.

\bibitem[Wang et~al.(2026{\natexlab{a}})Wang, Cai, Hoong, Walker, Kroiz,
  Vanica, Patil, Gavrilescu, Sipra, and Sankaran]{wang2026maxkernel}
S.~Wang, N.~Cai, C.~Hoong, J.~Walker, G.~Kroiz, G.~Vanica, D.~Patil,
  A.~Gavrilescu, H.~Sipra, and S.~Sankaran.
\newblock {MaxKernel}: Agentic kernel generation for {TPU}s.
\newblock \emph{arXiv preprint arXiv:2609.04523}, 2026{\natexlab{a}}.

\bibitem[Wang et~al.(2026{\natexlab{b}})Wang, Su, Zeng, Xu, Ren, Yang, Huang,
  He, Ma, Peng, Cheng, He, Chen, Wang, Du, and Shen]{wang2026thetaevolve}
Y.~Wang, S.-R. Su, Z.~Zeng, E.~Xu, L.~Ren, X.~Yang, Z.~Huang, X.~He, L.~Ma,
  B.~Peng, H.~Cheng, P.~He, W.~Chen, S.~Wang, S.~S. Du, and Y.~Shen.
\newblock Thetaevolve: Test-time learning on open problems.
\newblock In \emph{Forty-third International Conference on Machine Learning},
  2026{\natexlab{b}}.
\newblock URL \url{https://openreview.net/forum?id=XHP568OuIl}.

\bibitem[Wei et~al.(2025)Wei, Sun, Seenichamy, Song, Ouyang, Mirhoseini, Wang,
  and Aiken]{wei2025astra}
A.~Wei, T.~Sun, Y.~Seenichamy, H.~Song, A.~Ouyang, A.~Mirhoseini, K.~Wang, and
  A.~Aiken.
\newblock Astra: A multi-agent system for {GPU} kernel performance
  optimization.
\newblock In \emph{NeurIPS 2025 Fourth Workshop on Deep Learning for Code},
  2025.
\newblock URL \url{https://openreview.net/forum?id=IZKZIcPaHz}.

\bibitem[White et~al.(2021)White, Zela, Ru, Liu, and Hutter]{white2021powerful}
C.~White, A.~Zela, R.~Ru, Y.~Liu, and F.~Hutter.
\newblock How powerful are performance predictors in neural architecture
  search?
\newblock In \emph{Advances in Neural Information Processing Systems},
  volume~34, pages 28454--28469. Curran Associates, Inc., 2021.

\bibitem[Xing et~al.(2026)Xing, Zhai, Jiang, Dong, Wu, Ye, Ruan, Huang, Zhang,
  Yin, Bayyapu, Ceze, and Chen]{xing2026flashinfer}
S.~Xing, Y.~Zhai, A.~Jiang, Y.~Dong, Y.~Wu, Z.~Ye, C.~Ruan, Y.~Huang, Y.~Zhang,
  L.~Yin, A.~Bayyapu, L.~Ceze, and T.~Chen.
\newblock {FlashInfer-Bench}: Building the virtuous cycle for {AI}-driven {LLM}
  systems.
\newblock In \emph{Proceedings of Machine Learning and Systems}, volume~8,
  pages 2016--2064. MLSys, 2026.

\bibitem[Yan et~al.(2026)Yan, Peng, Coleman, Chen, Xie, Chen, He, Sachdeva, Ye,
  Wang, Wang, Chi, Pereira, Kang, Cheng, and Wang]{yan2026pacevolve}
M.~Yan, B.~Peng, B.~Coleman, Z.~Chen, Z.~Xie, S.~Chen, Z.~He, N.~Sachdeva,
  I.~Ye, W.~Wang, C.~Wang, E.~H. Chi, F.~Pereira, W.-C. Kang, D.~Z. Cheng, and
  B.~Wang.
\newblock {PACEvolve}: Enabling long-horizon progress-aware consistent
  evolution.
\newblock In \emph{ICLR 2026 Workshop on Lifelong Agents: Learning, Aligning,
  Evolving}, 2026.

\bibitem[Yang et~al.(2026)Yang, Fan, Pan, Xi, Wang, Sun, Keutzer, Han, Zaharia,
  Xu, and Stoica]{yang2026freetoken}
S.~Yang, X.~Fan, M.~Pan, H.~Xi, Z.~Wang, S.~Sun, K.~Keutzer, S.~Han,
  M.~Zaharia, C.~Xu, and I.~Stoica.
\newblock Freetoken: Efficient edge-native moe serving with bandwidth-adaptive
  execution.
\newblock \emph{arXiv preprint arXiv:2608.16157}, 2026.

\bibitem[Yuksekgonul et~al.(2026)Yuksekgonul, Koceja, Li, Bianchi, McCaleb,
  Wang, Kautz, Choi, Zou, Guestrin, and Sun]{yuksekgonul2026learning}
M.~Yuksekgonul, D.~Koceja, X.~Li, F.~Bianchi, J.~McCaleb, X.~Wang, J.~Kautz,
  Y.~Choi, J.~Zou, C.~Guestrin, and Y.~Sun.
\newblock Learning to discover at test time.
\newblock In \emph{Forty-third International Conference on Machine Learning},
  2026.
\newblock URL \url{https://openreview.net/forum?id=96zNuQrH9Y}.

\bibitem[Zadouri et~al.(2026)Zadouri, Hoehnerbach, Shah, Thakkar, and
  Dao]{zadouri2026flashattention}
T.~Zadouri, M.~Hoehnerbach, J.~Shah, V.~Thakkar, and T.~Dao.
\newblock {FlashAttention-4}: Algorithm and kernel pipelining co-design for
  asymmetric hardware scaling.
\newblock In \emph{Proceedings of Machine Learning and Systems}, volume~8,
  pages 912--926. MLSys, 2026.

\bibitem[Zou et~al.(2026)Zou, Zhu, and Dong]{kda2025}
D.~Zou, L.~Zhu, and Y.~Dong.
\newblock {KDA}: Kernel design agents for high-performance cuda kernels, 2026.
\newblock URL \url{https://github.com/NVlabs/kda}.
\newblock GitHub repository.

\end{thebibliography}
